\pdfoutput=1

\documentclass[11pt]{article}

\usepackage{acl}
\usepackage{times}
\usepackage{latexsym}
\usepackage{hyperref} 
\usepackage{tcolorbox}
\usepackage{subcaption}
\usepackage{array}
\usepackage{xcolor}
\usepackage{soul}
\definecolor{mygold}{HTML}{FFD243}
\definecolor{mygrey}{HTML}{CCD6D8}
\definecolor{mylightgreen}{HTML}{D1F694}
\definecolor{myblue}{HTML}{30C0F0}
\definecolor{myorange}{HTML}{F5AF22}
\definecolor{myred}{HTML}{FF576A}
\definecolor{darkred}{HTML}{91270F}
\definecolor{darkgreen}{HTML}{5E893E}
\sethlcolor{mylightgreen}
\usepackage{amssymb}
\usepackage{pifont}
\newcommand{\xmark}{\ding{55}}%
\newcommand{\cmark}{\textcolor{darkgreen}{\ding{51}}}

\usepackage[T1]{fontenc}

\usepackage[utf8]{inputenc}
\usepackage{graphicx}
\usepackage{amsmath,amssymb}
\usepackage[super]{nth}

\usepackage{microtype}
\usepackage{inconsolata}

\title{\textbf{\textcolor{myblue}{CO}\textcolor{myorange}{VE}}: COntext and VEracity prediction for out-of-context images}

\author{
Jonathan Tonglet$^{1,2,3}$, Gabriel Thiem$^{1}$, Iryna Gurevych$^{1}$
\\
        \textsuperscript{1}Ubiquitous Knowledge Processing Lab (UKP Lab), \\ Department of Computer Science 
and Hessian Center for AI (hessian.AI),\\ TU Darmstadt \\ 
\textsuperscript{2} Department of Electrical Engineering, KU Leuven\\
\textsuperscript{3} Department of Computer Science, KU Leuven\\
\href{https://www.ukp.tu-darmstadt.de}{www.ukp.tu-darmstadt.de}
}

\begin{document}
\maketitle
\begin{abstract}
Images taken out of their context are the most prevalent form of multimodal misinformation. Debunking them requires  (1) providing the true context of the image and (2) checking the veracity of the image's caption. However, existing automated fact-checking methods fail to tackle both objectives explicitly. In this work, we introduce \textbf{\textcolor{myblue}{CO}\textcolor{myorange}{VE}}, a new method that predicts first the true \textcolor{myblue}{COntext} of the image and then uses it to predict the \textcolor{myorange}{VEracity} of the caption. COVE beats the  SOTA context prediction model on all context items, often by more than five percentage points. It is competitive with the best veracity prediction models on synthetic data and outperforms them on real-world data, showing that it is beneficial to combine the two tasks sequentially. Finally, we conduct a human study that reveals that the predicted context is a reusable and interpretable artifact to verify new out-of-context captions for the same image. Our code and data are made available.\footnote{\href{https://github.com/UKPLab/naacl2025-cove}{github.com/UKPLab/naacl2025-cove}} 
\end{abstract}

\section{Introduction}
\label{sec:intro}

\begin{figure}[!ht]
    \centering
    \includegraphics[width=\linewidth]{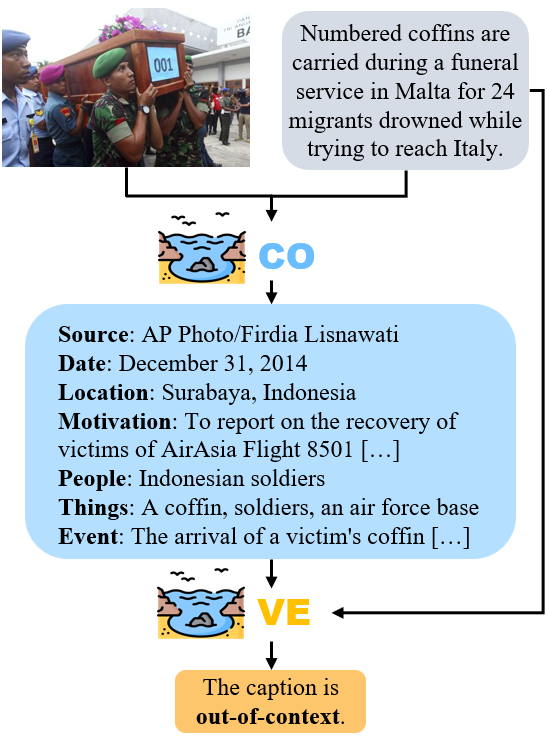}
    \caption{The two steps of \textbf{\textcolor{myblue}{CO}\textcolor{myorange}{VE}}: (1) Generating the true context of the image. (2) Predicting the veracity of a caption by comparing it with the generated context.}
    \label{fig:intro}
\end{figure}

An out-of-context (OOC) image is a form of multimodal misinformation where the caption misrepresents the image in one or several dimensions, including the date, location, or event depicted \citep{luo-etal-2021-newsclippings,dufour2024ammeba}. In 2023, more than 40\% of the visual misinformation verified by fact-checkers consisted of OOC images \citep{dufour2024ammeba}.
To debunk them, human fact-checkers follow two objectives: identifying the true context of the image, which usually takes the form of a fixed set of items \citep{silverman2013verification,urbani2020verifying,bellingcat2021,10017287,khan2024debunking},  and deciding on the veracity of the caption. Identifying the true context of the image first is often beneficial when checking the caption, as it may reveal inconsistencies with the image. However, it goes beyond that objective, as most context items are neither consistent nor inconsistent with the caption and provide more details than needed to verify the caption. 
In Figure \ref{fig:intro}, the context, shown in blue, reveals that the \textit{location} in the caption is inconsistent: Malta instead of Indonesia. It also discusses the \textit{source} and \textit{date}, which are absent from the caption, and provides a precise \textit{location} at the city level.

Many methods have been proposed to facilitate fact-checking (FC) of OOC images. However, they study either context prediction \citep{tonglet-etal-2024-image} or veracity prediction \citep{Abdelnabi_2022_CVPR,papadopoulos2023red,papadopoulos2024similarity,Qi_2024_CVPR}. They do not consider automating them sequentially, leveraging the comprehensive context to predict the veracity of the caption.

In this work, we introduce \textbf{\textcolor{myblue}{CO}\textcolor{myorange}{VE}}, the first method that both predicts a comprehensive  \textcolor{myblue}{COntext} for the image and a \textcolor{myorange}{VEracity}  label for the caption. The method consists of two steps, shown in Figure \ref{fig:intro}. First, the context is predicted as a set of seven items, three of which are first introduced in this work. Compared to prior work, COVE leverages a more diverse set of evidence to predict the context, including web search, knowledge bases, and the parametric knowledge of large language models (LLMs). Afterward, the caption is compared with the context to predict its veracity.
We summarize our contributions as follows. (1) Leveraging a more diverse set of evidence, COVE outperforms the context prediction SOTA  \citep{tonglet-etal-2024-image} for all context items, including three items introduced in this work, by 0.3 to 18.9 percentage points. (2)  COVE is competitive with the best models for veracity prediction on synthetic data and outperforms them on real-world data by up to 4.5 percentage points in Macro F1, highlighting the benefits of automating veracity prediction based on the predicted context. (3) Our experiments show that the predicted context is an interpretable artifact for human users, which they can reuse multiple times to verify new captions about the same image.

\section{Related work}

\textbf{Veracity prediction for OOC images}\\
Providing a veracity label for OOC images has received significant attention in automated fact-checking (AFC) research \citep{akhtar-etal-2023-multimodal}. It is a binary classification task with labels \{accurate, OOC\}. Synthetic datasets have been created by replacing entities in the true caption \citep{10.1145/3240508.3240707,10.1145/3372278.3390670}   or by mismatching image-caption pairs from news corpora  \citep{luo-etal-2021-newsclippings}. 
Smaller datasets based on real-world fact-checks have recently been proposed \citep{aneja2023cosmos,papadopoulos2024verite,pham2024ookpik}. Models that leverage external evidence outperform those that use only the image and the caption as input \citep{luo-etal-2021-newsclippings,zhang2023interpretable}. \citet{Abdelnabi_2022_CVPR} collects text evidence with reverse image search and image evidence by querying a search engine with the caption. The performance can further be improved by predicting the stance of the evidence towards the caption \citep{yuan-etal-2023-support} or ranking the most relevant evidence \citep{papadopoulos2023red}. Unlike prior methods, ECENet provides an explanation in the form of a summary of the most relevant text evidence \citep{10.1145/3581783.3612183}. Recent work leverages multimodal LLMs (MLLMs) with instruction-tuning \citep{Qi_2024_CVPR}, external tools \citep{braun2024defamedynamicevidencebasedfactchecking}, or multi-agent debate \citep{lakara2024madsherlockmultiagentdebatesoutofcontext}.  
\citet{papadopoulos2024similarity} showed that simple classifiers like random forest trained on top of the image, caption, and image and text evidence embeddings achieve SOTA performance. Their results show that veracity can often be predicted based on shallow heuristics, highlighting the need to assess the progress in the field from other perspectives \citep{papadopoulos2024similarity}, such as context prediction.\\

\noindent \textbf{Context prediction for OOC images}\\
Context prediction, or image contextualization, has received less attention in AFC. \citet{tonglet-etal-2024-image} formulate it as a question-answering (QA) task where each context item is predicted given the corresponding question and a set of evidence which may include the image, the caption, or pieces of information derived from them. \citet{tahmasebi2025verifyingcrossmodalentityconsistency} formulate it as a true/false classification task where candidate \textit{people}, \textit{locations}, and \textit{events} are verified with a MLLM. 5Pils is a real-world dataset \citep{tonglet-etal-2024-image} which  contains context labels based on human fact-checking practices \citep{urbani2020verifying,bellingcat2021,10017287,khan2024debunking}. \citet{tonglet-etal-2024-image} proposed a baseline for context prediction that retrieves text evidence with reverse image search and answers context questions with a (M)LLM and the image and the evidence as input.\\

In this work, we fill an important gap by introducing the first method that performs both tasks sequentially, leveraging the generated context for veracity prediction.

\section{COntext and VEracity (COVE)}
\begin{figure*}
    \centering
    \includegraphics[width=\linewidth]{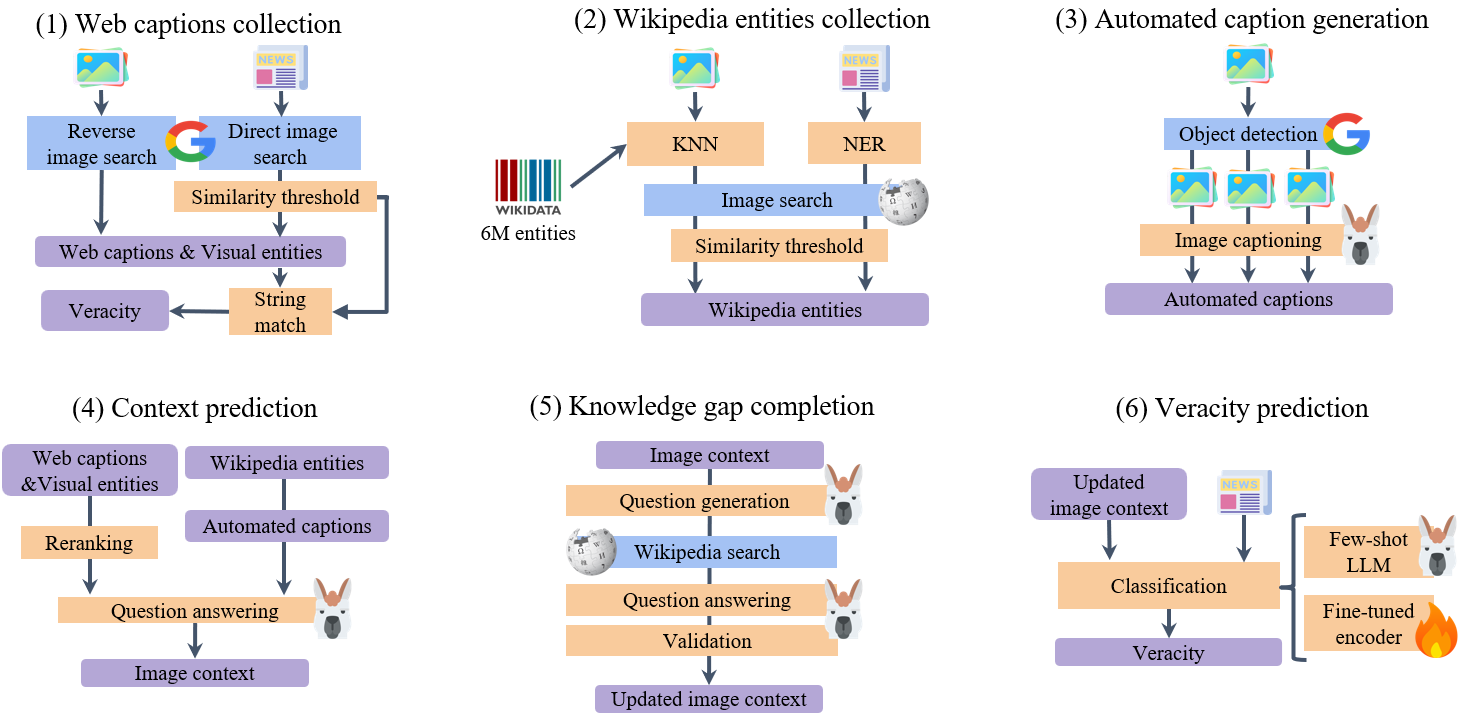}
    \caption{The architecture of \textbf{\textcolor{myblue}{CO}\textcolor{myorange}{VE}} consists of six steps. The first three are performed in parallel and consist of retrieving evidence. Step 4 predicts the context items in a QA setting. Step 5 updates missing items based on the existing ones and Wikipedia knowledge. Step 6 predicts the veracity of the caption based on the predicted context.}
    \label{fig:COVE}
\end{figure*}

COVE is a method to predict the true context of an image first and then the veracity of its caption. The overall architecture can be divided into six steps, as illustrated in Figure \ref{fig:COVE}. The first three steps are concerned with the collection of a diverse set of evidence, i.e., web captions and visual entities (§\ref{sec:web_cap}), Wikipedia entities (§\ref{sec:wiki_ent}), and automated captions (§\ref{sec:auto_cap}). Based on their similarity with web images and captions, the veracity of certain instances is already predicted during evidence retrieval (§\ref{sec:web_cap}). After collecting the evidence, a first version of the context is predicted with an LLM (§\ref{sec:context_QA}). Then,  some missing context items are updated by searching relevant Wikipedia passages (§\ref{sec:knowledge_QA}). Eventually, a model predicts the veracity of the caption based on the predicted context (§\ref{sec:veracity}).

\subsection{Web captions collection}
\label{sec:web_cap}
Following \citet{Abdelnabi_2022_CVPR}, we use reverse image search with the Google Vision API to retrieve web captions and visual entities associated with the same or partially matching images,\footnote{\href{https://cloud.google.com/vision/docs/detecting-web}{cloud.google.com/vision/docs/detecting-web}} and a custom Google search engine to retrieve relevant web images given the caption.\footnote{\href{https://programmablesearchengine.google.com/about/}{programmablesearchengine.google.com/about/}} We compute the cosine similarity between the CLIP \citep{pmlr-v139-radford21a} embeddings of the instance's image and the web images. For images with similarity above a threshold $t_{match}$, their attached caption is added to the set of web captions.

The web captions are sufficient to assign a veracity label to some instances, especially those for which the caption and/or the image come directly from newspapers. We apply the following three rules, which we refer to as veracity rules: (1) if the caption is an exact string match with one or more reverse image search web captions, the veracity is accurate, (2) if the similarity with a web image is above $t_{match}$, the veracity is accurate, (3) if the similarity with a web image is below threshold $t_{non\_match}$ and the attached web caption is an exact string match with the caption to verify, the veracity is OOC.

\subsection{Wikipedia entities collection}
\label{sec:wiki_ent}

\begin{figure}
    \centering
    \includegraphics[width=\linewidth]{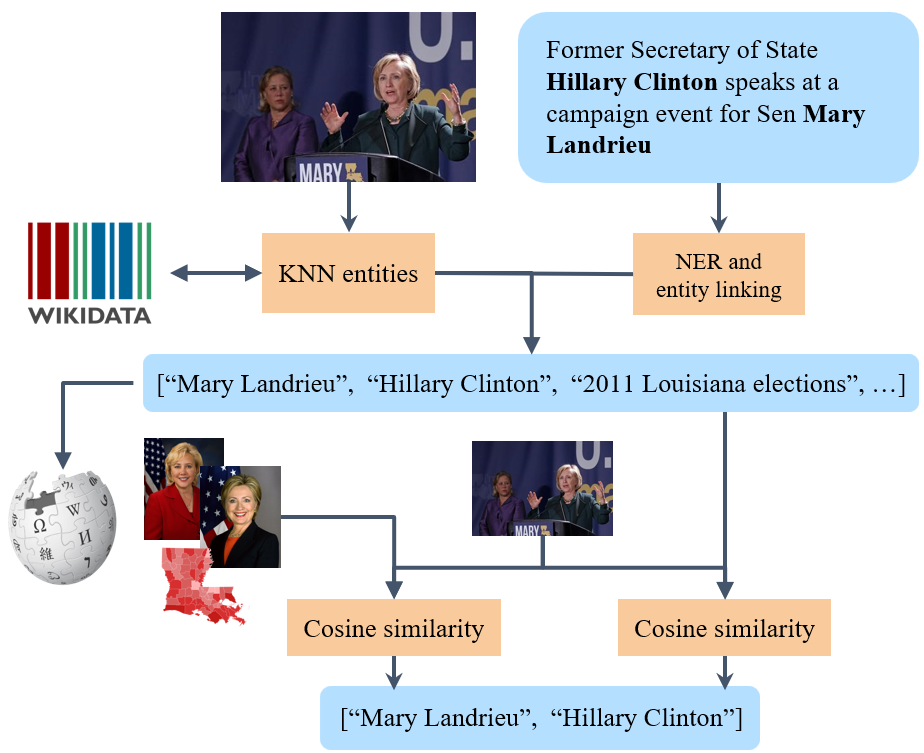}
    \caption{Wikipedia entities collection. The candidate set is composed of the entities in the caption and those that are most similar to the image. Candidates are retained if the similarity between the image and their name or their Wikipedia images passes a threshold.}
    \label{fig:entities}
\end{figure}

Images often contain entities, such as celebrities, products, and landmarks, that can be paired with a Wikipedia page \citep{10.1145/3372278.3390670}. Recognizing these entities and providing them as evidence can help predict the context. An example is illustrated in Figure \ref{fig:entities}. First, we collect a set of candidate entities: (1) we collect named entities from the caption and normalize them to match one or more Wikipedia entries using GENRE \citep{deautoregressive}, (2) we use the OVEN index \citep{hu2023open}, which contains 6 Million Wikidata entities, and retrieve the k nearest neighbors based on the cosine similarity between their CLIP text embeddings and the image embedding. For each candidate entity, if its embedding similarity with the image is superior to threshold $t_{wiki\_text}$, it is retained. Otherwise,  we scrape up to three images from the corresponding Wikipedia page and compute their embedding's cosine similarity with the image. If at least one similarity score is higher than $t_{wiki\_image}$, the entity is retained. 

\subsection{Automated caption generation}
\label{sec:auto_cap}

The third set of evidence is generated automatically with the MLLM LlavaNext \citep{10.5555/3666122.3667638,Liu_2024_CVPR}. One caption describing the entire image and another one describing the people are generated by providing the entire image as input. Then, following \citet{10.5555/3666122.3666162}, we detect objects in the image using the Google Vision API.\footnote{\href{https://cloud.google.com/vision/docs/object-localizer}{cloud.google.com/vision/docs/object-localizer}} We crop the image to its bounding boxes, and caption it with LlavaNext using a question specific to the object's category. Categories are manually defined with their questions in Appendix \ref{sec:object_categories}.

\subsection{Context prediction}
\label{sec:context_QA}
We predict seven context items: \{\textit{source}, \textit{date}, \textit{location}, \textit{motivation}, \textit{people}, \textit{things}, \textit{event}\}. The first four items were introduced in \citet{tonglet-etal-2024-image}, while the last ones are new to this work. 
The first \textit{l} web captions are ranked based on the number of named entities relevant to the context item that they contain, for example, the number of PERSON entities for the \textit{people} item.
The selected Wikipedia entities and the automated captions are merged into two captions: one for everything that relates to \textit{people} in the image and the other one for all other pieces of information. 
All the collected evidence is provided as input with a question to the LLM Llama 3 \citep{dubey2024llama3}  to predict a context item. When the evidence does not contain the answer, the model is prompted to output ``Unknown''. Each item, its associated question, and relevant named entities are discussed in Appendix \ref{sec:items} .

\subsection{Knowledge gap completion}
\label{sec:knowledge_QA}

\begin{figure}
    \centering
    \includegraphics[width=\linewidth]{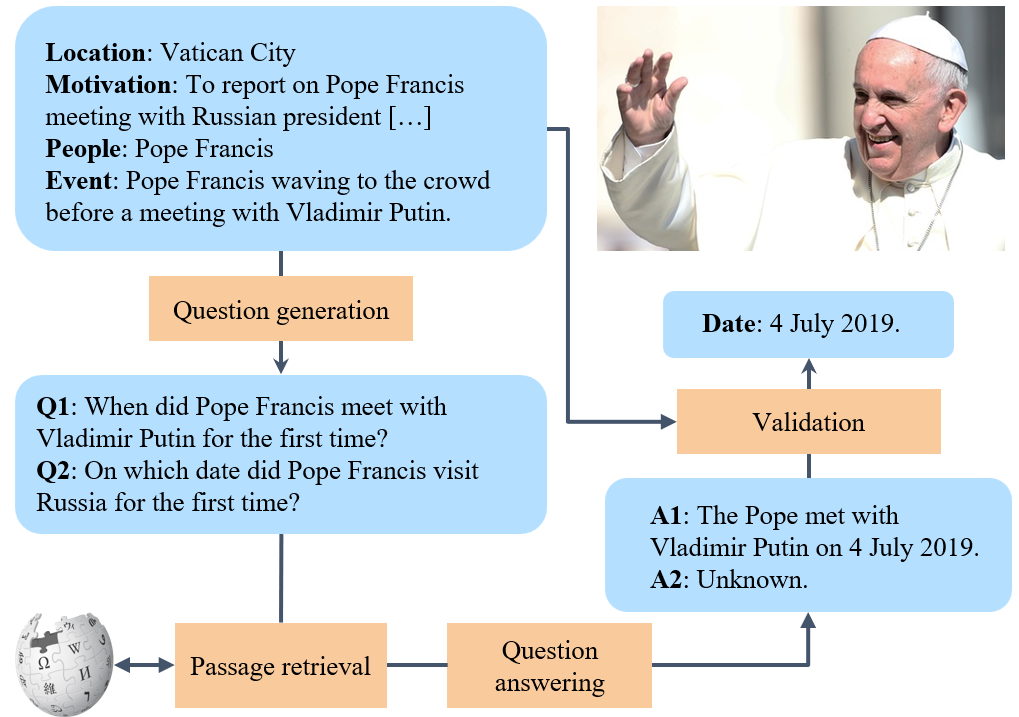}
    \caption{Knowledge gap completion. Questions are generated based on the predicted context and answered with Wikipedia passages. If the answers are relevant, the context is updated.}
    \label{fig:knowledge}
\end{figure}
The \textit{date} and \textit{location} can often be missing after step 4. However, these two items can be derived from other context items and world knowledge. Following QACheck \citep{pan-etal-2023-qacheck}, we use a sequence of three modules: question generation, question answering,  and validation, illustrated in Figure \ref{fig:knowledge}. If the answer to \textit{date} is Unknown, and \textit{location} + \textit{people}, \textit{event}, or \textit{motivation} are available, we provide the currently known context items as input to Llama 3 and ask it to generate up to three questions that could help predict the date of the image. For each question, relevant Wikipedia passages are retrieved with WikiChat and Colbert \citep{10.1145/3397271.3401075,semnani-etal-2023-wikichat} and provided as input to Llama 3 to provide an answer. Finally, Llama 3 predicts the \textit{date}, if it can be determined, based on the existing context and the generated QA pairs. 
For \textit{location}, we apply the same pipeline if the \textit{date} + \textit{people}, \textit{event}, or \textit{motivation} are available. More examples of knowledge gap completions are shown in Appendix \ref{sec:knowledge_gap}.

\subsection{Veracity prediction}
\label{sec:veracity}

This step takes the predicted context and the caption as input and is performed only for captions that could not automatically be verified based on the web evidence (§\ref{sec:web_cap}). We consider two different models. (1) A frozen Llama 3 with few-shot demonstrations as input. Each demonstration's output starts with an explanation of the inconsistencies, followed by the predicted veracity, which is either ``accurate'',   ``OOC'', ``Unknown, probably [label]'', which counts as a prediction of the corresponding label, or ``Unknown'', which is mapped to the majority predicted label. This flexibility in the output answers was empirically found to help Llama 3 in ambiguous cases. (2) A DebertaV3 \citep{hedebertav3} model fine-tuned on (predicted context, ground truth veracity) training pairs.

\section{Experiments}

\subsection{Datasets}

 \textbf{NewsCLIPpings}  is a synthetic dataset for veracity prediction \citep{luo-etal-2021-newsclippings}. Images and their captions are selected from Visual News \citep{liu-etal-2021-visual}. OOC pairs are created by mismatching images and captions using various measures of semantic similarity. Following prior works, we use the ``merged-balanced'' split, which contains 71,072, 7,024, and 7,264 instances in the train, validation, and test splits, respectively. Within a split, each caption appears twice, once as accurate and once as OOC. NewsCLIPpings does not contain ground truth context items. Hence, we create them by decomposing the accurate caption in a set of context items with Llama 3, as explained in Appendix \ref{sec:GT_context}.

\textbf{5Pils-OOC} is a real-world test set containing 624 images, each paired with two captions, one accurate and the other OOC, for a total of 1248 instances. We construct 5Pils-OOC as a subset of 5Pils \citep{tonglet-etal-2024-image}. The images and OOC captions have been fact-checked by human experts from three organizations: Factly, Pesacheck, and 211Check. The images come from both Western and non-Western contexts, in particular, India and Ethiopia. As 5Pils does not contain accurate captions, we generate them automatically using GPT4 \citep{openai2023gpt4} based on the ground truth context items. Furthermore, 5Pils does not contain ground truth labels for context items \textit{people}, \textit{things}, and \textit{event}. They are derived from the accurate caption, as explained in Appendix \ref{sec:GT_context}. The creation of 5Pils-OOC from 5Pils is detailed in Appendix \ref{sec:5Pils_OOC}.

\subsection{Context metrics}
We report one metric per context item. Additional evaluations with all metrics introduced in \citet{tonglet-etal-2024-image} are provided in Appendix \ref{sec:metrics}.

 \textit{Source}, \textit{motivation}, \textit{things}, and \textit{event}  are evaluated with Meteor \citep{banerjee-lavie-2005-meteor}.

\textit{Date} predictions are mapped to timestamps. We use $\Delta$, which is inversely proportional to the distance in years with the ground truth.

 \textit{Location} predictions are mapped to coordinates using GeoNames.\footnote{\href{https://www.geonames.org/}{geonames.org}} We use Coordinates $\Delta$ (CO$\Delta$), which is inversely proportional to the distance in thousand kilometers with the ground truth.

 \textit{People} expects sets of named entities as predictions, evaluated with the Macro F1-score (F1).

\subsection{Veracity metrics}
We use four metrics for veracity prediction: the accuracy A, the recall over accurate samples R$_{ACC}$, the recall over OOC  samples R$_{OOC}$, and F1.

\subsection{Baselines}

For context prediction, we compare COVE with the 5Pils baseline \citet{tonglet-etal-2024-image}, which provides the image and web captions from reverse image search as input to a MLLM, LlavaNext in our case, asking one question per context item.

We compare COVE against three SOTA models for veracity prediction of OOC images, which all rely on external evidence.
\textbf{RED-DOT} \citep{papadopoulos2023red} and \textbf{AITR} \citep{papadopoulos2024similarity} use transformer architectures trained on top of the CLIP embeddings of the image, the caption, and text and image evidence. Furthermore, they filter the evidence to keep only the most relevant one from each modality.
\textbf{SNIFFER} \citep{Qi_2024_CVPR} predicts veracity based on two signals: (1)  the detection of inconsistencies between the caption, the image, and the visual entities, using a fine-tuned InstructBLIP \citep{10.5555/3666122.3668264}, and (2) the detection of inconsistencies between the caption and text evidence using a frozen Vicuna \citep{chiang2023vicuna}, which we replace here by Llama 3 for a fair comparison. We also evaluate veracity prediction with Llama 3 based on the context items predicted by the 5Pils baseline. Finally, we report as an upper bound the COVE veracity results using the ground truth context items.

\subsection{Implementation details}
All hyperparameters are tuned, and ablations are performed on a random sample of 1500 instances from the NewsCLIPpings validation set.
DebertaV3 is fine-tuned on a subset of the NewsCLIPpings train set containing 5000 instances.
Inference with Llama 3 is done in a few-shot setting with 4 to 8 demonstrations, which have been selected and labeled by hand from a random sample of 100 NewsCLIPpings train instances. Model versions and the hyperparameters are listed in Appendix \ref{sec:implementation}. Appendix \ref{sec:prompts} reports the prompts for Llama 3.

The same set of web evidence is used for all methods. Following RED-DOT, AITR, and SNIFFER, we only use the caption field of the web evidence for NewsCLIPpings. For 5Pils-OOC, we use the title field because few web evidence provides a caption. This is more restrictive than the setup of \citet{tonglet-etal-2024-image} and limits context prediction by ignoring important webpage fields like the publication date. However, it ensures a fair comparison with the veracity prediction baselines. Unlike NewsCLIPpings, the web evidence of 5Pils-OOC are multilingual, including texts in Amharic, Arabic, Hindi, and Telugu.

\subsection{Main results}
\label{sec:results}

\begin{table}
    \centering
    \resizebox{\columnwidth}{!}{ %
    \begin{tabular}{cccccccc}
    \hline 
        & Source & Date & Loc. & Mot. & People & Things & Event  \\
        & (M) & ($\Delta$) & (CO$\Delta$) & (M) & (F1) & (M) & (M) \\
    \hline
        & \multicolumn{7}{c}{\textbf{NewsCLIPpings}}\\
    \hline
      5Pils baseline & 3.4 & 28.9 & 32.7 & 1.5 &  39.9 &  9.9 & 15.4 \\
      \textbf{\textcolor{myblue}{CO}\textcolor{myorange}{VE}}   & \textbf{8.1} & \textbf{41.5} & \textbf{51.6}  & \textbf{11.6} & \textbf{49.0} & \textbf{10.9} & \textbf{22.4} \\
        \hline
        & \multicolumn{7}{c}{\textbf{5Pils-OOC}}\\
    \hline
    5Pils baseline  & 0.3 &    1.8 & 21.8 & 3.0 & 12.8 & 4.9 & 4.8  \\ 
      \textbf{\textcolor{myblue}{CO}\textcolor{myorange}{VE}}       & \textbf{0.6} & \textbf{7.0} & \textbf{28.9}& \textbf{15.1} & \textbf{20.5} &  \textbf{7.2} &  \textbf{9.4} \\
    \hline
     \end{tabular}}
    \caption{Context prediction results on the test sets (\%). The best scores are marked in \textbf{bold}. Loc. and Mot. are the location and the motivation, respectively.}
    \label{tab:context_results}
\end{table}

\begin{table}
    \centering
    \resizebox{\columnwidth}{!}{ %
    \begin{tabular}{cccccccccc}
       \hline
       & \multicolumn{4}{c}{\textbf{NewsCLIPpings}}  & & \multicolumn{4}{c}{\textbf{5Pils-OOC}} \\
         & (A) & (R$_{ACC}$) & (R$_{OOC}$) & (F1) & & (A) & (R$_{ACC}$) & (R$_{OOC}$) & (F1) \\
         \hline
    RED-DOT \includegraphics[height=1em]{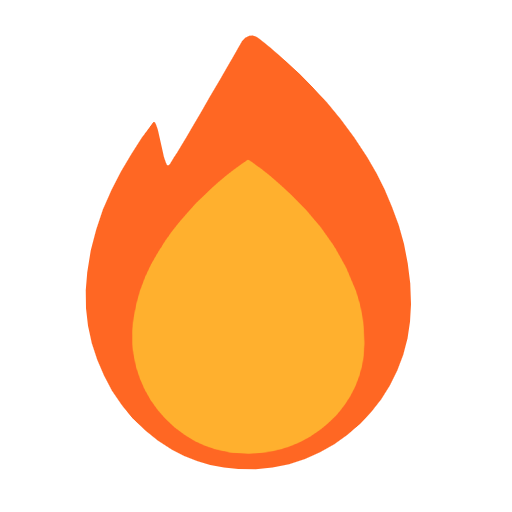}   & 90.3  & 87.3 & 93.3 & 90.3 & & 46.8 & 42.8 & 50.8 & 46.7 \\
    AITR \includegraphics[height=1em]{figures/fire.png}  & \textbf{93.5} & \textbf{94.8} & \textbf{92.1} & \textbf{93.5} & & 52.6 & 81.4 & 23.9 & 48.4 \\
    SNIFFER \includegraphics[height=1em]{figures/fire.png} \includegraphics[height=1em]{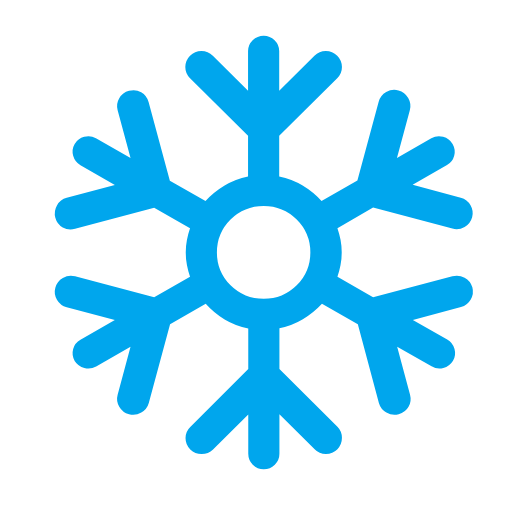}  & 88.4  & 92.3 & 84.4 & 88.3 &  & 56.3 & \textbf{86.4} &  26.1 & 51.9 \\
    5Pils baseline - LLama 3 \includegraphics[height=1em]{figures/snowflake.png} & 77.2 & 74.6 & 79.8 &  77.2 &  & 55.7 &  42.8 & 68.6 &  54.9 \\
    \hline
    & \multicolumn{9}{c}{\textbf{\textcolor{myblue}{CO}\textcolor{myorange}{VE}} with predicted context}\\
    \hline
    DebertaV3 \includegraphics[height=1em]{figures/fire.png} & 87.9  & 84.0 & 91.8 & 87.9 & & 56.7 & 47.6 & 65.9 & \textbf{56.4}  \\
    Llama 3 \includegraphics[height=1em]{figures/snowflake.png} & 86.7  & 83.5 & 89.9 & 86.7 & & \textbf{58.2} & 33.0 & \textbf{83.3} &  55.3 \\
    \hline 
    & \multicolumn{9}{c}{\textbf{\textcolor{myblue}{CO}\textcolor{myorange}{VE}} with ground truth context}\\
    \hline
    DebertaV3 \includegraphics[height=1em]{figures/fire.png} & 94.1 & 99.2 & 88.9 & 94.0 & & 80.7 & 100.0 & 61.4 & 79.9\\
    Llama 3 \includegraphics[height=1em]{figures/snowflake.png} & 95.1 & 97.1 & 93.1 & 94.4 &  & 95.9 & 99.7 & 92.1 & 95.3 \\
    \hline
    \end{tabular}}
    \caption{Veracity prediction results on the test sets (\%). \includegraphics[height=1em]{figures/fire.png} and \includegraphics[height=1em]{figures/snowflake.png}  indicate trained models and frozen models used in a few-shot setting, respectively. The best scores without ground truth are marked in \textbf{bold}.}
    \label{tab:veracity_results}
\end{table}

\begin{table*}
    \centering
    \resizebox{\textwidth}{!}{ %
    \begin{tabular}{ccccccccccccccccccc}
    \hline 
       \rotatebox{90}{Web captions \:  \: \: \: } & 
       \rotatebox{90}{Visual entities  \: \: }  & \rotatebox{90}{Wikipedia entities  \: \: } 
       &  \rotatebox{90}{Automated captions \: \:}  & \rotatebox{90}{Knowledge gap  \: \: } & \rotatebox{90}{Context prediction  \: \: }  & \rotatebox{90}{Veracity prediction  \: \: } & \rotatebox{90}{Veracity rules  \: \: }    & Source & Date & Location & Motivation & People & Things & Event & \multicolumn{4}{c}{Veracity} \\
       \cline{16-19}
        &   & & &   & & &  & (M) & ($\Delta$) & (CO$\Delta$) & (M) & (F1) & (M) & (M) &   (A) & (R$_{ACC}$) & (R$_{OOC}$) & (F1)\\
    \hline
         \cmark    &  \cmark   &  \cmark     &   \cmark   &  \cmark   & \cmark    & \cmark    & \cmark   & 11.2  & \textbf{48.2} &  \textbf{55.7}   &  \textbf{11.2} & 53.0 & 10.0 &  \textbf{22.2} & \textbf{88.3} & 86.8 & \textbf{89.8}  & \textbf{88.3} \\
      \cmark   & \xmark    &  \xmark     &  \xmark    & \xmark       &  \xmark &  \xmark & \cmark   & - & - & - & - & - & - & - & 53.1 & 69.8 & 35.5 &  44.4\\
     \cmark    &  \cmark   &  \cmark   & \cmark    & \cmark     & \cmark   & \cmark    & \xmark  &  11.2  & \textbf{48.2} &  \textbf{55.7}   &  \textbf{11.2} & 53.0 & 10.0 &  \textbf{22.2} & 84.2 & 80.2 & 88.4 & 84.2 \\
      \cmark    &  \cmark   &  \cmark   & \cmark    & \cmark     &
    \xmark   & \cmark    & \cmark  &  - & - & - & - & - & - & -& 78.2 & \textbf{94.9} & 60.7 & 77.4 \\
    \cmark      &     \cmark     &  \xmark  & \xmark   &  \xmark & \cmark & \cmark  & \cmark   & 11.2 & 43.7  & 49.6 & 10.8 &  45.1 & \textbf{11.1} & 17.7 & 85.1 & 94.1 & 75.7 & 78.6 \\
      \cmark &   \xmark     &  \cmark   &  \cmark   &  \xmark   & \cmark & \cmark   & \xmark  & \textbf{14.1} & \textbf{48.2} & 54.5 & 10.7 & \textbf{53.6} & 10.7 & 21.0  & \textbf{88.3} & 88.5 & 88.1  &  \textbf{88.3}\\
       \xmark   & \cmark &    \cmark    &  \cmark   &  \xmark & \cmark  & \cmark    &   \cmark  & 1.5 & 9.6 & 14.8  & 9.3 & 28.1  & 6.9 & 9.5 & 77.9 & 73.2 & 82.8  & 77.9 \\
       \xmark & \xmark      &  \cmark    &  \cmark    & \xmark & \cmark  & \cmark   &\xmark & 1.5 &  8.2 & 10.1 &  9.1 & 27.5 & 7.4 & 8.5 & 77.1 &  73.0 & 81.4  & 77.1 \\
       \xmark & \xmark      &  \cmark    & \cmark     &  \cmark  & \cmark  & \cmark   &\xmark & 1.5  & 8.8 &  11.8  &  9.1 & 27.5 & 7.4 & 8.5 & 76.8 & 71.1  &  82.8 & 76.8 \\

        \hline
    \end{tabular}}
    \caption{Ablation results on the NewsCLIPpings validation subset (\%). Veracity metrics are reported for few-shot Llama 3. \cmark and \xmark \: indicate included and removed components, respectively. The best scores are marked in \textbf{bold}.}
    \label{tab:ablation}
\end{table*}

 \noindent \textbf{Context prediction}\\
Thanks to its more diverse evidence set, COVE outperforms the 5Pils baseline \citep{tonglet-etal-2024-image} on all context items and both datasets, as shown in Table \ref{tab:context_results}. The performance increases from 1.0 to 18.9 on NewsCLIPpings, and from 0.3 to 12.1 percentage points on 5Pils-OOC. For \textit{date}, \textit{location}, \textit{motivation},  and \textit{event}, COVE is not only more accurate but also abstains less often from answering than the baseline. We attribute these improvements to the larger and more diverse evidence set and the knowledge gap completion step for \textit{date} and \textit{location}. In particular, on 5Pils-OOC,  including the knowledge gap completion more than doubles the COVE \textit{date} scores, from 3.3 to 7.0\%, and increases the \textit{location} scores by 6.1 percentage points. Thanks to the Wikipedia entities retrieval, COVE achieves up to 9.1 percentage points higher F1 than the baseline for \textit{people}. \textit{Source} relies a lot on web captions, which are also part of the evidence set in the baseline, resulting in limited improvements.

With both methods, better contextualization is achieved on NewsCLIPpings. The largest drops in performance when moving to 5Pils-OOC are observed for \textit{source}, \textit{date}, \textit{location}, and \textit{people}. The following properties of the images in NewsCLIPpings explain this: (1) they are of higher quality, (2) they all originate from news articles and the retrieved web captions contain more information and are less noisy, impacting \textit{source} the most, (3) most of them are set in a Western context, on which MLLMs tend to perform better \citep{ananthram2024see}, while most of 5Pils-OOC images are set in East Africa and South Asia. This affects \textit{location} the most, which is often estimated based on the automated captions if no web captions are available.\\

\noindent \textbf{Veracity prediction}\\
The veracity results of COVE are competitive with the baselines on NewsCLIPpings, as shown in Table \ref{tab:veracity_results}, suffering mainly from a low R$_{ACC}$ while achieving near SOTA R$_{OOC}$. COVE with DebertaV3 achieves slightly better performance, which we attribute to the fine-tuning on NewsCLIPpings train instances. 

All methods suffer from lower results when switching from synthetic data to the 5Pils-OOC real-world data. However, COVE becomes the strongest method, achieving better accuracy, R$_{OOC}$, and F1 than the baselines. While COVE with Llama 3 achieves the best accuracy, COVE with DebertaV3 is the best method in terms of F1. In comparison, RED-DOT performs worse than random,  and AITR and SNIFFER have very low R$_{OOC}$, as low as one OOC caption on four. RED-DOT and AITR always predict the OOC label for instances that have web captions, highlighting their reliance on shallow heuristics that do not generalize beyond NewsCLIPpings \citep{papadopoulos2024similarity}. While fine-tuned on synthetic data like the baselines, DebertaV3 generalizes better to 5Pils-OOC, which we attribute to leveraging a comprehensive and structured context as input.

 Providing the context items predicted by the 5Pils baseline as input to Llama 3 results in a drop in performance, in particular on NewsCLIPpings where the difference in context prediction performance between the 5Pils baseline and COVE is the largest.
 Furthermore, we validate that the COVE results using the ground truth context lead to near-perfect R$_{ACC}$, both with DebertaV3 and Llama 3. This is expected because the ground truth context items are obtained by decomposing accurate captions. On NewsCLIPpings, COVE achieves F1 of 94.4 and 94.0 \% with Llama 3 and DebertaV3, respectively, higher than all baselines. These results confirm that predicting a comprehensive and accurate context and providing it as input ensures high performance for veracity prediction. While Llama 3 with ground truth context achieves a similar performance on 5Pils-OOC, DebertaV3 faces a large drop in  R$_{OOC}$. We assume that DebertaV3 suffers like the baselines, although to a lower extent, from being trained on synthetic OOC data, which follow a different distribution than the real-world data of 5Pils-OOC.

\subsection{Ablation study}

We report ablation results on the validation subset in Table \ref{tab:ablation}, using Llama 3 for veracity prediction. By relying solely on the veracity rules for web captions (§\ref{sec:web_cap}), an accuracy of 53.1\% is obtained, and more than a third of the OOC captions can be detected. This happens without predicting any context element.

Removing veracity rules for web captions deteriorates R$_{ACC}$, and R$_{OOC}$ to a lower extent. Removing context prediction, providing instead the raw evidence as input for veracity prediction, decreases  R$_{OOC}$ by 29.1 percentage points. This shows that using the predicted context as input is a decisive factor favoring R$_{OOC}$ over R$_{ACC}$.

The accuracy decreases, and a large imbalance in favor of R$_{ACC}$ appears when using only web captions and visual entities. Furthermore, the performance on several context items decreases too, by up to 7.9 percentage points for \textit{people}, highlighting the need to consider a more diverse set of evidence. For \textit{source} and \textit{people}, the best results are achieved without visual entities, indicating that they may contain irrelevant information and conflict with the other evidence. All context and veracity metrics experience a large drop when removing web captions. In particular,  \textit{source} relies the most on web captions. Removing both the visual entities and web captions decreases 
 performance further. However, the F1 for \textit{people} remains more than half of what would be obtained with web results, thanks to the Wikipedia entities collection step. Furthermore, COVE still detects more than 80\% of the OOC captions, and the F1 for veracity is only 1.5 percentage points lower than the one obtained with web captions and visual entities only. In the absence of web results, knowledge gap completion can improve \textit{date} and \textit{location} scores but only to a small extent.

\subsection{Human study}

\begin{table}
    \centering
    \resizebox{\columnwidth}{!}{ %
    \begin{tabular}{cccc}
         \hline
         & A & F1 & $\kappa$ \\
         \hline
       Group 1 - no AFC artifact  &   38.9 & 38.7 & 23.0 \\
       Group 1 - SNIFFER artifact & 60.0 & 58.3 & 44.7 \\
       Group 2 - no AFC artifact  & 34.4   & 33.8 & 34.7\\
       Group 2 - COVE artifact & \textbf{85.6} & \textbf{83.0} & \textbf{60.8} \\
       \hline
    \end{tabular}}
    \caption{Human study results (\%). $\kappa$ is Fleiss' $\kappa$. The best scores are marked in  \textbf{bold}. }
    \label{tab:human_results}
\end{table}

\begin{figure}
    \centering
    \includegraphics[width=\linewidth]{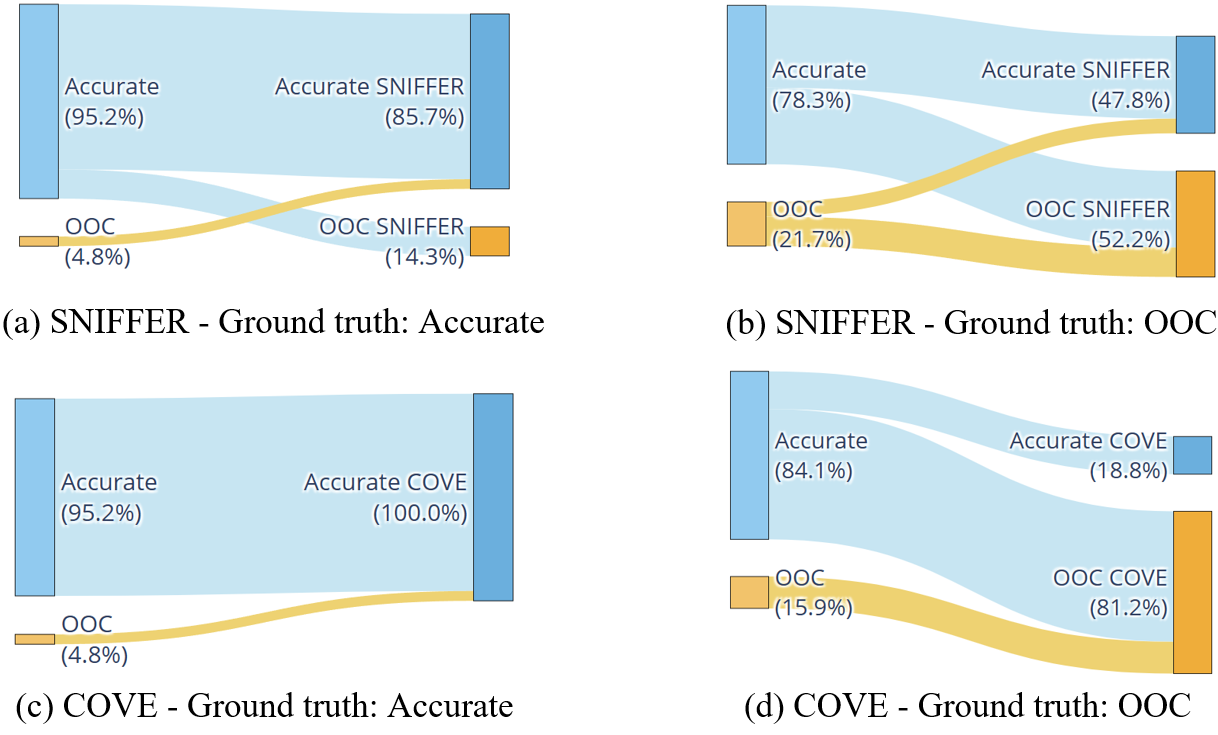}
    \caption{Change in veracity prediction before and after seeing SNIFFER (top) or COVE (bottom) artifacts, for accurate (left) and OOC (right) captions.}
    \label{fig:human}
\end{figure}

Several OOC captions can appear over time for an image, each of them miscaptioning the image in a different way. We assess to which extent the output of AFC models constitutes a reusable artifact for humans to verify new captions about the same image. To the best of our knowledge, this is the first study of this important property of AFC artifacts. We compare two artifacts: the explanations generated by SNIFFER, which focuses on inconsistency detection, and the context predicted by COVE.

The study is split into two phases. (1) Given an image and an old OOC caption correctly detected by the AFC model, the human annotator has to classify three new captions about the same image. (2) The annotator classifies again the same new captions, this time given the AFC artifact generated for the old caption. We use the majority vote of the annotators in each phase as the label assigned to a new caption. Following \citet{Qi_2024_CVPR}, we ask participants to provide confidence levels: not, somewhat, or highly confident.

We collect 30 OOC instances from the NewsCLIPpings test set and construct three new captions for each. This results in 22 and 68 accurate and OOC captions, respectively. Appendix \ref{sec:human_statistics} explains the creation of the new captions. Annotators instructions and examples are shown in Appendix \ref{sec:human_example}.

We recruited 6 students and provided the SNIFFER explanations to half of them and the COVE context to the others. Table \ref{tab:human_results} shows that both groups perform poorly without AFC artifacts. This means that knowing that the old caption is OOC does not inform the annotators much about the veracity of the new ones. Hence, models that only provide a veracity score, like RED-DOT and AITR, have limited purpose beyond a specific (image-caption) pair. COVE contexts are more useful artifacts than  SNIFFER explanations to verify new captions, as demonstrated by the larger increase in accuracy, Macro F1, and inter-annotator agreement (Fleiss' $\kappa$ (\citep{fleiss1971measuring}) in the second phase. We attribute this to COVE's more comprehensive approach, predicting context items even if they are not relevant to verify the old caption, e.g., the caption does not contain a date, but COVE still predicts the true date. On the other hand, SNIFFER often explains the minimal set of inconsistent elements to detect an OOC caption, which might be insufficient to verify new captions.

Figure \ref{fig:human} shows the change in predictions between the two phases of the study. In the first phase, both groups label the majority of captions as accurate. Upon observing the COVE artifact, annotators successfully identify all accurate captions, and the number of detected OOC captions increases by more than 65 percentage points. By comparison, the SNIFFER artifact results in a 30.5 percentage points increase in detected OOC captions. Furthermore, observing the COVE artifact exclusively improves predictions by correcting errors from the first phase, whereas observing the SNIFFER artifact introduces new errors, shifting some correct predictions to incorrect ones.

This experiment confirms that COVE provides a reusable artifact to verify new captions, thanks to its comprehensive context prediction.
We report aggregated confidence levels in Appendix \ref{sec:confidence_level}.

\subsection{Error analysis}

We manually analyze random samples of 200 instances from each test set and report the distribution of errors in Table \ref{tab:errors}.
There are five error categories.

(1) Incorrect context items are predicted, which leads to veracity errors, in particular for accurate captions. In 5Pils-OOC, more than half of these errors stem from irrelevant Wikipedia entities detected by CLIP. This issue is especially pronounced in African and South Asian contexts, which are prevalent in 5Pils-OOC. In contrast, this problem does not arise in NewsCLIPpings, where most images have a Western context.  Irrelevant or inaccurate web captions further contribute to context prediction errors.

(2) Context items are missing. This prevents Llama 3 from verifying all the atomic facts in the caption. The most frequently missing items are the \textit{date}, \textit{location}, and \textit{event}. While \textit{people} is important in NewsCLIPpings, its absence has a smaller impact in 5Pils-OOC. This is because misrepresenting the individuals in the image is less frequent in real-world misinformation.  

(3) Only one error in NewsCLIPpings is due to an instance passing the $t_{match}$ veracity rule while actually being OOC. (4) In a few cases, the context items are correct and sufficient to predict the veracity, but Llama 3 makes a reasoning error. (5) We found some errors in the ground truth, where the caption is not a description of the image. Therefore, the instance is not suitable for veracity prediction.

This analysis highlights directions for future work, including a better similarity matching of Wikipedia entities for images with non-Western contexts and a better ranking of web captions based on their relevance. We provide three error examples in Appendix \ref{sec:qualitative_analysis}.

\begin{table}\resizebox{\columnwidth}{!}{ %
\begin{tabular}{  m{11em} c c  }
    \hline
        & NewsCLIPpings & 5Pils-OOC  \\
    \hline
    Incorrect items & 20.5 & 57.1  \\
    $\hookrightarrow$  Web captions error & 6.8 & 14.3  \\
    $\hookrightarrow$ Wikipedia entities error & 0.0 & 33.3 \\
    $\hookrightarrow$ Auto. captions error & 9.1 & 7.1 \\
    $\hookrightarrow$ Knowledge gap error &  4.6 & 2.4 \\
    \hline
    Missing items   & 54.6 & 36.9  \\
    $\hookrightarrow$ Missing date & 15.9 & 26.2 \\
    $\hookrightarrow$ Missing location & 29.5 & 27.4 \\
    $\hookrightarrow$ Missing people & 15.9 & 3.6 \\
    $\hookrightarrow$ Missing things & 9.1 & 1.2 \\
    $\hookrightarrow$ Missing event & 31.8 & 20.2 \\
    \hline
    Rule-based error & 2.3 & 0.0  \\ 
    Llama 3 veracity error & 6.8 & 1.2\\
    Error in ground truth & 15.9 & 4.8  \\
    \hline
\end{tabular}}
  \caption{Error distribution (\%) on the test sets with few-shot Llama 3. $\hookrightarrow$ indicates a sub-category.}
\label{tab:errors}
\end{table}

\section{Conclusion}

We propose \textbf{\textcolor{myblue}{CO}\textcolor{myorange}{VE}}, a new method to combat OOC misinformation that predicts first the true context of an image and leverages it to predict the veracity of its caption. COVE outperforms the SOTA on context prediction for all context items while being competitive with the best veracity models, even outperforming them on real-world data by up to 4.5 percentage points in Macro F1, showcasing the benefits of sequentially performing context and veracity prediction. Furthermore, our human study shows that the predicted context is a useful and reusable artifact for human users to verify new captions for the same image. 

\section{Limitations}

We identify four limitations to this work.

(1) Similar to other QA methods in AFC \citep{pan-etal-2023-fact,pan-etal-2023-qacheck,khaliq-etal-2024-ragar,10.5555/3666122.3668964}, COVE requires several (M)LLM inference steps which are computationally expensive. However, the context predictions and automated captions can be generated in batches, improving inference speed. Furthermore, some steps of COVE can be removed depending on the computational budget available. For example,  knowledge gap completion requires many LLM calls while being less critical to the pipeline than web caption and Wikipedia entity collection.

(2) We did not consider the use of closed-source LLMs like GPT4 \citep{openai2023gpt4} for context and veracity prediction. However, this work does not have the objective to provide a performance comparison between open and closed-source LLMs. COVE is LLM-agnostic, and we expect its performance to improve with LLMs that perform better on standard benchmarks and leaderboards. 

(3) The context items of NewsCLIPpings and parts of the items of 5Pils-OOC are weakly labeled by decomposing accurate captions. However, there is no guarantee that the ground truth provides the most comprehensive summary of the context. In some cases, the context predictions might be correct and more detailed than the ground truth, e.g., predicting the correct location at the town level, while the ground truth is at the country level. The predictions are slightly penalized in the evaluation metrics when providing more precise answers despite them being correct. This limitation is also present in the 5Pils dataset \citep{tonglet-etal-2024-image}.  

(4) Some of the retrieved web captions may contain misinformation themselves or come from unreliable sources. However, there is no filtering mechanism implemented to remove web captions collected from unreliable websites. In Appendix \ref{sec:filtering}, we discuss a simple filtering mechanism that keeps web evidence from a manually defined list of reliable web domains, with a small negative impact on performance. Future work should consider the inclusion of models designed for evaluating the reliability of web evidence \citep{10.1145/3643491.3660278} and their sources \citep{schlichtkrull-2024-generating}.

\section{Ethics statement}

\textbf{Intended uses}
COVE addresses the important societal problem of multimodal misinformation by contextualizing images and detecting OOC captions.
While AFC methods have made significant progress over the years, they are still prone to errors. In particular, the context predicted by COVE is still far from reaching sufficient quality, and as shown in Table \ref{tab:veracity_results}, only three OOC captions out of four are detected on real-world data. Given the high negative impact of labeling misinformation as true information, and vice-versa, AFC methods like COVE  should only be used as complementary assistants to a human FC expert.\\

\noindent \textbf{Misuse potential} 
COVE and other AFC methods show promising results in detecting misinformation. As a result, they could also be used in an adversarial setup by malicious actors to craft misinformation that is harder to detect by AFC methods and human fact-checkers. Nevertheless, we believe that the benefits of supporting the work of human fact-checkers with partial automation outweigh the risks caused by this malicious adversarial setup.

\section*{Acknowledgements}

This work has been funded by the LOEWE initiative (Hesse, Germany) within the emergenCITY center (Grant Number: LOEWE/1/12/519/03/05.001(0016)/72) and by the German Federal Ministry of Education and Research and the Hessian Ministry of Higher Education, Research, Science and the Arts within their joint support of the National Research Center for Applied Cybersecurity ATHENE. We gratefully acknowledge the support of Microsoft with a grant for access to OpenAI GPT models via the Azure cloud (Accelerate Foundation Model Academic Research). Figures 1 and 3 have been designed using resources from Flaticon.com. We want to express our gratitude to Xiuying Chen, Hiba Arnaout, and Fengyu Cai for their insightful comments on a draft of this paper.

\bibliography{anthology,custom}

\begin{thebibliography}{53}
\providecommand{\natexlab}[1]{#1}

\bibitem[{Abdelnabi et~al.(2022)Abdelnabi, Hasan, and Fritz}]{Abdelnabi_2022_CVPR}
Sahar Abdelnabi, Rakibul Hasan, and Mario Fritz. 2022.
\newblock \href {https://doi.org/10.1109/CVPR52688.2022.01452} {Open-domain, content-based, multi-modal fact-checking of out-of-context images via online resources}.
\newblock In \emph{{IEEE/CVF} Conference on Computer Vision and Pattern Recognition, {CVPR} 2022, New Orleans, LA, USA, June 18-24, 2022}, pages 14920--14929. {IEEE}.

\bibitem[{Akhtar et~al.(2023)Akhtar, Schlichtkrull, Guo, Cocarascu, Simperl, and Vlachos}]{akhtar-etal-2023-multimodal}
Mubashara Akhtar, Michael Schlichtkrull, Zhijiang Guo, Oana Cocarascu, Elena Simperl, and Andreas Vlachos. 2023.
\newblock \href {https://doi.org/10.18653/v1/2023.findings-emnlp.361} {Multimodal automated fact-checking: A survey}.
\newblock In \emph{Findings of the Association for Computational Linguistics: EMNLP 2023}, pages 5430--5448, Singapore. Association for Computational Linguistics.

\bibitem[{Ananthram et~al.(2024)Ananthram, Stengel-Eskin, Vondrick, Bansal, and McKeown}]{ananthram2024see}
Amith Ananthram, Elias Stengel-Eskin, Carl Vondrick, Mohit Bansal, and Kathleen McKeown. 2024.
\newblock \href {https://arxiv.org/abs/2406.11665} {See it from my perspective: Diagnosing the western cultural bias of large vision-language models in image understanding}.
\newblock \emph{ArXiv preprint}, abs/2406.11665.

\bibitem[{Aneja et~al.(2023)Aneja, Bregler, and Nie{\ss}ner}]{aneja2023cosmos}
Shivangi Aneja, Chris Bregler, and Matthias Nie{\ss}ner. 2023.
\newblock \href {https://doi.org/10.1609/AAAI.V37I12.26648} {{COSMOS:} catching out-of-context image misuse using self-supervised learning}.
\newblock In \emph{Thirty-Seventh {AAAI} Conference on Artificial Intelligence, {AAAI} 2023, Thirty-Fifth Conference on Innovative Applications of Artificial Intelligence, {IAAI} 2023, Thirteenth Symposium on Educational Advances in Artificial Intelligence, {EAAI} 2023, Washington, DC, USA, February 7-14, 2023}, pages 14084--14092. {AAAI} Press.

\bibitem[{Banerjee and Lavie(2005)}]{banerjee-lavie-2005-meteor}
Satanjeev Banerjee and Alon Lavie. 2005.
\newblock \href {https://aclanthology.org/W05-0909} {{METEOR}: An automatic metric for {MT} evaluation with improved correlation with human judgments}.
\newblock In \emph{Proceedings of the {ACL} Workshop on Intrinsic and Extrinsic Evaluation Measures for Machine Translation and/or Summarization}, pages 65--72, Ann Arbor, Michigan. Association for Computational Linguistics.

\bibitem[{Braun et~al.(2024)Braun, Rothermel, Rohrbach, and Rohrbach}]{braun2024defamedynamicevidencebasedfactchecking}
Tobias Braun, Mark Rothermel, Marcus Rohrbach, and Anna Rohrbach. 2024.
\newblock \href {https://arxiv.org/abs/2412.10510} {Defame: Dynamic evidence-based fact-checking with multimodal experts}.
\newblock \emph{ArXiv preprint}, abs/2412.10510.

\bibitem[{Cao et~al.(2021)Cao, Izacard, Riedel, and Petroni}]{deautoregressive}
Nicola~De Cao, Gautier Izacard, Sebastian Riedel, and Fabio Petroni. 2021.
\newblock \href {https://openreview.net/forum?id=5k8F6UU39V} {Autoregressive entity retrieval}.
\newblock In \emph{9th International Conference on Learning Representations, {ICLR} 2021, Virtual Event, Austria, May 3-7, 2021}. OpenReview.net.

\bibitem[{Chiang et~al.(2023)Chiang, Li, Lin, Sheng, Wu, Zhang, Zheng, Zhuang, Zhuang, Gonzalez, Stoica, and Xing}]{chiang2023vicuna}
Wei-Lin Chiang, Zhuohan Li, Zi~Lin, Ying Sheng, Zhanghao Wu, Hao Zhang, Lianmin Zheng, Siyuan Zhuang, Yonghao Zhuang, Joseph~E. Gonzalez, Ion Stoica, and Eric~P. Xing. 2023.
\newblock \href {https://lmsys.org/blog/2023-03-30-vicuna/} {Vicuna: An open-source chatbot impressing gpt-4 with 90\%* chatgpt quality}.

\bibitem[{Chrysidis et~al.(2024)Chrysidis, Papadopoulos, Papadopoulos, and Petrantonakis}]{10.1145/3643491.3660278}
Zacharias Chrysidis, Stefanos-Iordanis Papadopoulos, Symeon Papadopoulos, and Panagiotis Petrantonakis. 2024.
\newblock \href {https://doi.org/10.1145/3643491.3660278} {Credible, unreliable or leaked?: Evidence verification for enhanced automated fact-checking}.
\newblock In \emph{Proceedings of the 3rd ACM International Workshop on Multimedia AI against Disinformation}, MAD '24, page 73–81, New York, NY, USA. Association for Computing Machinery.

\bibitem[{Dai et~al.(2023)Dai, Li, Li, Tiong, Zhao, Wang, Li, Fung, and Hoi}]{10.5555/3666122.3668264}
Wenliang Dai, Junnan Li, Dongxu Li, Anthony Meng~Huat Tiong, Junqi Zhao, Weisheng Wang, Boyang Li, Pascale Fung, and Steven C.~H. Hoi. 2023.
\newblock \href {http://papers.nips.cc/paper\_files/paper/2023/hash/9a6a435e75419a836fe47ab6793623e6-Abstract-Conference.html} {Instructblip: Towards general-purpose vision-language models with instruction tuning}.
\newblock In \emph{Advances in Neural Information Processing Systems 36: Annual Conference on Neural Information Processing Systems 2023, NeurIPS 2023, New Orleans, LA, USA, December 10 - 16, 2023}.

\bibitem[{Douze et~al.(2024)Douze, Guzhva, Deng, Johnson, Szilvasy, Mazar{\'e}, Lomeli, Hosseini, and J{\'e}gou}]{douze2024faiss}
Matthijs Douze, Alexandr Guzhva, Chengqi Deng, Jeff Johnson, Gergely Szilvasy, Pierre-Emmanuel Mazar{\'e}, Maria Lomeli, Lucas Hosseini, and Herv{\'e} J{\'e}gou. 2024.
\newblock \href {https://arxiv.org/abs/2401.08281} {The faiss library}.
\newblock \emph{ArXiv preprint}, abs/2401.08281.

\bibitem[{Dufour et~al.(2024)Dufour, Pathak, Samangouei, Hariri, Deshetti, Dudfield, Guess, Escayola, Tran, Babakar et~al.}]{dufour2024ammeba}
Nicholas Dufour, Arkanath Pathak, Pouya Samangouei, Nikki Hariri, Shashi Deshetti, Andrew Dudfield, Christopher Guess, Pablo~Hern{\'a}ndez Escayola, Bobby Tran, Mevan Babakar, et~al. 2024.
\newblock \href {https://arxiv.org/abs/2405.11697} {Ammeba: A large-scale survey and dataset of media-based misinformation in-the-wild}.
\newblock \emph{ArXiv preprint}, abs/2405.11697.

\bibitem[{Fleiss(1971)}]{fleiss1971measuring}
Joseph~L Fleiss. 1971.
\newblock \href {https://doi.org/10.1037/h0031619} {Measuring nominal scale agreement among many raters.}
\newblock \emph{Psychological bulletin}, 76(5):378--382.

\bibitem[{He et~al.(2023)He, Gao, and Chen}]{hedebertav3}
Pengcheng He, Jianfeng Gao, and Weizhu Chen. 2023.
\newblock \href {https://openreview.net/pdf?id=sE7-XhLxHA} {Debertav3: Improving deberta using electra-style pre-training with gradient-disentangled embedding sharing}.
\newblock In \emph{The Eleventh International Conference on Learning Representations, {ICLR} 2023, Kigali, Rwanda, May 1-5, 2023}. OpenReview.net.

\bibitem[{Hu et~al.(2023{\natexlab{a}})Hu, Luan, Chen, Khandelwal, Joshi, Lee, Toutanova, and Chang}]{hu2023open}
Hexiang Hu, Yi~Luan, Yang Chen, Urvashi Khandelwal, Mandar Joshi, Kenton Lee, Kristina Toutanova, and Ming{-}Wei Chang. 2023{\natexlab{a}}.
\newblock \href {https://doi.org/10.1109/ICCV51070.2023.01108} {Open-domain visual entity recognition: Towards recognizing millions of wikipedia entities}.
\newblock In \emph{{IEEE/CVF} International Conference on Computer Vision, {ICCV} 2023, Paris, France, October 1-6, 2023}, pages 12031--12041. {IEEE}.

\bibitem[{Hu et~al.(2023{\natexlab{b}})Hu, Iscen, Sun, Chang, Sun, Ross, Schmid, and Fathi}]{10.5555/3666122.3666162}
Ziniu Hu, Ahmet Iscen, Chen Sun, Kai{-}Wei Chang, Yizhou Sun, David Ross, Cordelia Schmid, and Alireza Fathi. 2023{\natexlab{b}}.
\newblock \href {http://papers.nips.cc/paper\_files/paper/2023/hash/029df12a9363313c3e41047844ecad94-Abstract-Conference.html} {{AVIS:} autonomous visual information seeking with large language model agent}.
\newblock In \emph{Advances in Neural Information Processing Systems 36: Annual Conference on Neural Information Processing Systems 2023, NeurIPS 2023, New Orleans, LA, USA, December 10 - 16, 2023}.

\bibitem[{Khaliq et~al.(2024)Khaliq, Chang, Ma, Pflugfelder, and Mileti{\'c}}]{khaliq-etal-2024-ragar}
Mohammed~Abdul Khaliq, Paul Yu-Chun Chang, Mingyang Ma, Bernhard Pflugfelder, and Filip Mileti{\'c}. 2024.
\newblock \href {https://doi.org/10.18653/v1/2024.fever-1.29} {{RAGAR}, your falsehood radar: {RAG}-augmented reasoning for political fact-checking using multimodal large language models}.
\newblock In \emph{Proceedings of the Seventh Fact Extraction and VERification Workshop (FEVER)}, pages 280--296, Miami, Florida, USA. Association for Computational Linguistics.

\bibitem[{Khan et~al.(2024)Khan, Dierickx, Furuly, Vold, Tahseen, Linden, and Dang-Nguyen}]{khan2024debunking}
Sohail~Ahmed Khan, Laurence Dierickx, Jan-Gunnar Furuly, Henrik~Brattli Vold, Rano Tahseen, Carl-Gustav Linden, and Duc-Tien Dang-Nguyen. 2024.
\newblock \href {https://doi.org/10.1002/asi.24970} {Debunking war information disorder: A case study in assessing the use of multimedia verification tools}.
\newblock \emph{Journal of the Association for Information Science and Technology}.

\bibitem[{Khan et~al.(2023)Khan, Sheikhi, Opdahl, Rabbi, Stoppel, Trattner, and Dang-Nguyen}]{10017287}
Sohail~Ahmed Khan, Ghazaal Sheikhi, Andreas~L. Opdahl, Fazle Rabbi, Sergej Stoppel, Christoph Trattner, and Duc-Tien Dang-Nguyen. 2023.
\newblock \href {https://doi.org/10.1109/ACCESS.2023.3236993} {Visual user-generated content verification in journalism: An overview}.
\newblock \emph{IEEE Access}, 11:6748--6769.

\bibitem[{Khattab and Zaharia(2020)}]{10.1145/3397271.3401075}
Omar Khattab and Matei Zaharia. 2020.
\newblock \href {https://doi.org/10.1145/3397271.3401075} {Colbert: Efficient and effective passage search via contextualized late interaction over {BERT}}.
\newblock In \emph{Proceedings of the 43rd International {ACM} {SIGIR} conference on research and development in Information Retrieval, {SIGIR} 2020, Virtual Event, China, July 25-30, 2020}, pages 39--48. {ACM}.

\bibitem[{Kwon et~al.(2023)Kwon, Li, Zhuang, Sheng, Zheng, Yu, Gonzalez, Zhang, and Stoica}]{10.1145/3600006.3613165}
Woosuk Kwon, Zhuohan Li, Siyuan Zhuang, Ying Sheng, Lianmin Zheng, Cody~Hao Yu, Joseph Gonzalez, Hao Zhang, and Ion Stoica. 2023.
\newblock \href {https://doi.org/10.1145/3600006.3613165} {Efficient memory management for large language model serving with pagedattention}.
\newblock In \emph{Proceedings of the 29th Symposium on Operating Systems Principles}, SOSP '23, page 611–626, New York, NY, USA. Association for Computing Machinery.

\bibitem[{Lakara et~al.(2024)Lakara, Sock, Rupprecht, Torr, Collomosse, and de~Witt}]{lakara2024madsherlockmultiagentdebatesoutofcontext}
Kumud Lakara, Juil Sock, Christian Rupprecht, Philip Torr, John Collomosse, and Christian~Schroeder de~Witt. 2024.
\newblock \href {https://arxiv.org/abs/2410.20140} {Mad-sherlock: Multi-agent debates for out-of-context misinformation detection}.
\newblock \emph{ArXiv preprint}, abs/2410.20140.

\bibitem[{Lin(2004)}]{lin-2004-rouge}
Chin-Yew Lin. 2004.
\newblock \href {https://aclanthology.org/W04-1013} {{ROUGE}: A package for automatic evaluation of summaries}.
\newblock In \emph{Text Summarization Branches Out}, pages 74--81, Barcelona, Spain. Association for Computational Linguistics.

\bibitem[{Liu et~al.(2021)Liu, Wang, Wang, and Ordonez}]{liu-etal-2021-visual}
Fuxiao Liu, Yinghan Wang, Tianlu Wang, and Vicente Ordonez. 2021.
\newblock \href {https://doi.org/10.18653/v1/2021.emnlp-main.542} {Visual news: Benchmark and challenges in news image captioning}.
\newblock In \emph{Proceedings of the 2021 Conference on Empirical Methods in Natural Language Processing}, pages 6761--6771, Online and Punta Cana, Dominican Republic. Association for Computational Linguistics.

\bibitem[{Liu et~al.(2024)Liu, Li, Li, and Lee}]{Liu_2024_CVPR}
Haotian Liu, Chunyuan Li, Yuheng Li, and Yong~Jae Lee. 2024.
\newblock \href {https://openaccess.thecvf.com/content/CVPR2024/html/Liu_Improved_Baselines_with_Visual_Instruction_Tuning_CVPR_2024_paper.html} {Improved baselines with visual instruction tuning}.
\newblock In \emph{Proceedings of the IEEE/CVF Conference on Computer Vision and Pattern Recognition (CVPR)}, pages 26296--26306.

\bibitem[{Liu et~al.(2023)Liu, Li, Wu, and Lee}]{10.5555/3666122.3667638}
Haotian Liu, Chunyuan Li, Qingyang Wu, and Yong~Jae Lee. 2023.
\newblock \href {http://papers.nips.cc/paper\_files/paper/2023/hash/6dcf277ea32ce3288914faf369fe6de0-Abstract-Conference.html} {Visual instruction tuning}.
\newblock In \emph{Advances in Neural Information Processing Systems 36: Annual Conference on Neural Information Processing Systems 2023, NeurIPS 2023, New Orleans, LA, USA, December 10 - 16, 2023}.

\bibitem[{Luo et~al.(2021)Luo, Darrell, and Rohrbach}]{luo-etal-2021-newsclippings}
Grace Luo, Trevor Darrell, and Anna Rohrbach. 2021.
\newblock \href {https://doi.org/10.18653/v1/2021.emnlp-main.545} {{N}ews{CLIP}pings: {A}utomatic {G}eneration of {O}ut-of-{C}ontext {M}ultimodal {M}edia}.
\newblock In \emph{Proceedings of the 2021 Conference on Empirical Methods in Natural Language Processing}, pages 6801--6817, Online and Punta Cana, Dominican Republic. Association for Computational Linguistics.

\bibitem[{MetaAI(2024)}]{dubey2024llama3}
MetaAI. 2024.
\newblock \href {https://arxiv.org/abs/2407.21783} {The llama 3 herd of models}.
\newblock \emph{ArXiv preprint}, abs/2407.21783.

\bibitem[{Mossou and Higgins(2021)}]{bellingcat2021}
Annique Mossou and Ross Higgins. 2021.
\newblock \href {https://www.bellingcat.com/resources/2021/11/01/a-beginners-guide-to-social-media-verification/} {A beginner's guide to social media verification}.
\newblock Accessed: 2023-09-15.

\bibitem[{M\"{u}ller-Budack et~al.(2020)M\"{u}ller-Budack, Theiner, Diering, Idahl, and Ewerth}]{10.1145/3372278.3390670}
Eric M\"{u}ller-Budack, Jonas Theiner, Sebastian Diering, Maximilian Idahl, and Ralph Ewerth. 2020.
\newblock \href {https://doi.org/10.1145/3372278.3390670} {Multimodal analytics for real-world news using measures of cross-modal entity consistency}.
\newblock In \emph{Proceedings of the 2020 International Conference on Multimedia Retrieval}, ICMR '20, page 16–25, New York, NY, USA. Association for Computing Machinery.

\bibitem[{OpenAI(2023)}]{openai2023gpt4}
OpenAI. 2023.
\newblock \href {https://arxiv.org/abs/2303.08774} {Gpt-4 technical report}.
\newblock Technical report, OpenAI.

\bibitem[{Pan et~al.(2023{\natexlab{a}})Pan, Lu, Kan, and Nakov}]{pan-etal-2023-qacheck}
Liangming Pan, Xinyuan Lu, Min-Yen Kan, and Preslav Nakov. 2023{\natexlab{a}}.
\newblock \href {https://doi.org/10.18653/v1/2023.emnlp-demo.23} {{QAC}heck: A demonstration system for question-guided multi-hop fact-checking}.
\newblock In \emph{Proceedings of the 2023 Conference on Empirical Methods in Natural Language Processing: System Demonstrations}, pages 264--273, Singapore. Association for Computational Linguistics.

\bibitem[{Pan et~al.(2023{\natexlab{b}})Pan, Wu, Lu, Luu, Wang, Kan, and Nakov}]{pan-etal-2023-fact}
Liangming Pan, Xiaobao Wu, Xinyuan Lu, Anh~Tuan Luu, William~Yang Wang, Min-Yen Kan, and Preslav Nakov. 2023{\natexlab{b}}.
\newblock \href {https://doi.org/10.18653/v1/2023.acl-long.386} {Fact-checking complex claims with program-guided reasoning}.
\newblock In \emph{Proceedings of the 61st Annual Meeting of the Association for Computational Linguistics (Volume 1: Long Papers)}, pages 6981--7004, Toronto, Canada. Association for Computational Linguistics.

\bibitem[{Papadopoulos et~al.(2023)Papadopoulos, Koutlis, Papadopoulos, and Petrantonakis}]{papadopoulos2023red}
Stefanos-Iordanis Papadopoulos, Christos Koutlis, Symeon Papadopoulos, and Panagiotis~C Petrantonakis. 2023.
\newblock \href {https://arxiv.org/abs/2311.09939} {Red-dot: Multimodal fact-checking via relevant evidence detection}.
\newblock \emph{ArXiv preprint}, abs/2311.09939.

\bibitem[{Papadopoulos et~al.(2024{\natexlab{a}})Papadopoulos, Koutlis, Papadopoulos, and Petrantonakis}]{papadopoulos2024similarity}
Stefanos-Iordanis Papadopoulos, Christos Koutlis, Symeon Papadopoulos, and Panagiotis~C Petrantonakis. 2024{\natexlab{a}}.
\newblock \href {https://arxiv.org/abs/2407.13488} {Similarity over factuality: Are we making progress on multimodal out-of-context misinformation detection?}
\newblock \emph{ArXiv preprint}, abs/2407.13488.

\bibitem[{Papadopoulos et~al.(2024{\natexlab{b}})Papadopoulos, Koutlis, Papadopoulos, and Petrantonakis}]{papadopoulos2024verite}
Stefanos-Iordanis Papadopoulos, Christos Koutlis, Symeon Papadopoulos, and Panagiotis~C Petrantonakis. 2024{\natexlab{b}}.
\newblock \href {https://doi.org/10.1007/s13735-023-00312-6} {Verite: a robust benchmark for multimodal misinformation detection accounting for unimodal bias}.
\newblock \emph{International Journal of Multimedia Information Retrieval}, 13(1):4.

\bibitem[{Pham et~al.(2024)Pham, Nguyen-Nhat, Dinh, Le, Nguyen, Tran, Tran, and Dang-Nguyen}]{pham2024ookpik}
Kha-Luan Pham, Minh-Khoi Nguyen-Nhat, Anh-Huy Dinh, Quang-Tri Le, Manh-Thien Nguyen, Anh-Duy Tran, Minh-Triet Tran, and Duc-Tien Dang-Nguyen. 2024.
\newblock \href {https://doi.org/10.1007/978-3-031-53302-0_10} {Ookpik- a collection of out-of-context image-caption pairs}.
\newblock In \emph{MultiMedia Modeling}, pages 132--144, Cham. Springer Nature Switzerland.

\bibitem[{Qi et~al.(2024)Qi, Yan, Hsu, and Lee}]{Qi_2024_CVPR}
Peng Qi, Zehong Yan, Wynne Hsu, and Mong~Li Lee. 2024.
\newblock \href {https://openaccess.thecvf.com/content/CVPR2024/html/Qi_SNIFFER_Multimodal_Large_Language_Model_for_Explainable_Out-of-Context_Misinformation_Detection_CVPR_2024_paper.html} {Sniffer: Multimodal large language model for explainable out-of-context misinformation detection}.
\newblock In \emph{Proceedings of the IEEE/CVF Conference on Computer Vision and Pattern Recognition (CVPR)}, pages 13052--13062.

\bibitem[{Radford et~al.(2021)Radford, Kim, Hallacy, Ramesh, Goh, Agarwal, Sastry, Askell, Mishkin, Clark, Krueger, and Sutskever}]{pmlr-v139-radford21a}
Alec Radford, Jong~Wook Kim, Chris Hallacy, Aditya Ramesh, Gabriel Goh, Sandhini Agarwal, Girish Sastry, Amanda Askell, Pamela Mishkin, Jack Clark, Gretchen Krueger, and Ilya Sutskever. 2021.
\newblock \href {http://proceedings.mlr.press/v139/radford21a.html} {Learning transferable visual models from natural language supervision}.
\newblock In \emph{Proceedings of the 38th International Conference on Machine Learning, {ICML} 2021, 18-24 July 2021, Virtual Event}, volume 139 of \emph{Proceedings of Machine Learning Research}, pages 8748--8763. {PMLR}.

\bibitem[{Sabir et~al.(2018)Sabir, AbdAlmageed, Wu, and Natarajan}]{10.1145/3240508.3240707}
Ekraam Sabir, Wael AbdAlmageed, Yue Wu, and Prem Natarajan. 2018.
\newblock \href {https://doi.org/10.1145/3240508.3240707} {Deep multimodal image-repurposing detection}.
\newblock In \emph{2018 {ACM} Multimedia Conference on Multimedia Conference, {MM} 2018, Seoul, Republic of Korea, October 22-26, 2018}, pages 1337--1345.

\bibitem[{Schlichtkrull et~al.(2023)Schlichtkrull, Guo, and Vlachos}]{10.5555/3666122.3668964}
Michael Schlichtkrull, Zhijiang Guo, and Andreas Vlachos. 2023.
\newblock \href {http://papers.nips.cc/paper\_files/paper/2023/hash/cd86a30526cd1aff61d6f89f107634e4-Abstract-Datasets\_and\_Benchmarks.html} {Averitec: {A} dataset for real-world claim verification with evidence from the web}.
\newblock In \emph{Advances in Neural Information Processing Systems 36: Annual Conference on Neural Information Processing Systems 2023, NeurIPS 2023, New Orleans, LA, USA, December 10 - 16, 2023}.

\bibitem[{Schlichtkrull(2024)}]{schlichtkrull-2024-generating}
Michael~Sejr Schlichtkrull. 2024.
\newblock \href {https://doi.org/10.18653/v1/2024.findings-emnlp.283} {Generating media background checks for automated source critical reasoning}.
\newblock In \emph{Findings of the Association for Computational Linguistics: EMNLP 2024}, pages 4927--4947, Miami, Florida, USA. Association for Computational Linguistics.

\bibitem[{Semnani et~al.(2023)Semnani, Yao, Zhang, and Lam}]{semnani-etal-2023-wikichat}
Sina Semnani, Violet Yao, Heidi Zhang, and Monica Lam. 2023.
\newblock \href {https://doi.org/10.18653/v1/2023.findings-emnlp.157} {{W}iki{C}hat: Stopping the hallucination of large language model chatbots by few-shot grounding on {W}ikipedia}.
\newblock In \emph{Findings of the Association for Computational Linguistics: EMNLP 2023}, pages 2387--2413, Singapore. Association for Computational Linguistics.

\bibitem[{Silverman(2013)}]{silverman2013verification}
Craig Silverman. 2013.
\newblock \href {https://verificationhandbook.com/} {Verification handbook}.
\newblock Accessed: 2023-09-15.

\bibitem[{Tahmasebi et~al.(2025)Tahmasebi, Müller-Budack, and Ewerth}]{tahmasebi2025verifyingcrossmodalentityconsistency}
Sahar Tahmasebi, Eric Müller-Budack, and Ralph Ewerth. 2025.
\newblock \href {https://arxiv.org/abs/2501.11403} {Verifying cross-modal entity consistency in news using vision-language models}.
\newblock \emph{ArXiv preprint}, abs/2501.11403.

\bibitem[{Tonglet et~al.(2024)Tonglet, Moens, and Gurevych}]{tonglet-etal-2024-image}
Jonathan Tonglet, Marie-Francine Moens, and Iryna Gurevych. 2024.
\newblock \href {https://doi.org/10.18653/v1/2024.emnlp-main.448} {{\textquotedblleft}image, tell me your story!{\textquotedblright} predicting the original meta-context of visual misinformation}.
\newblock In \emph{Proceedings of the 2024 Conference on Empirical Methods in Natural Language Processing}, pages 7845--7864, Miami, Florida, USA. Association for Computational Linguistics.

\bibitem[{Urbani(2020)}]{urbani2020verifying}
Shaydanay Urbani. 2020.
\newblock \href {https://firstdraftnews.org/long-form-article/verifying-online-information/} {Verifying online information}.
\newblock Accessed: 2023-09-15.

\bibitem[{Wolf et~al.(2020)Wolf, Debut, Sanh, Chaumond, Delangue, Moi, Cistac, Rault, Louf, Funtowicz, Davison, Shleifer, von Platen, Ma, Jernite, Plu, Xu, Le~Scao, Gugger, Drame, Lhoest, and Rush}]{wolf-etal-2020-transformers}
Thomas Wolf, Lysandre Debut, Victor Sanh, Julien Chaumond, Clement Delangue, Anthony Moi, Pierric Cistac, Tim Rault, Remi Louf, Morgan Funtowicz, Joe Davison, Sam Shleifer, Patrick von Platen, Clara Ma, Yacine Jernite, Julien Plu, Canwen Xu, Teven Le~Scao, Sylvain Gugger, Mariama Drame, Quentin Lhoest, and Alexander Rush. 2020.
\newblock \href {https://doi.org/10.18653/v1/2020.emnlp-demos.6} {Transformers: State-of-the-art natural language processing}.
\newblock In \emph{Proceedings of the 2020 Conference on Empirical Methods in Natural Language Processing: System Demonstrations}, pages 38--45, Online. Association for Computational Linguistics.

\bibitem[{Yuan et~al.(2023)Yuan, Guo, Qiu, Huang, and Li}]{yuan-etal-2023-support}
Xin Yuan, Jie Guo, Weidong Qiu, Zheng Huang, and Shujun Li. 2023.
\newblock \href {https://doi.org/10.18653/v1/2023.emnlp-main.259} {Support or refute: Analyzing the stance of evidence to detect out-of-context mis- and disinformation}.
\newblock In \emph{Proceedings of the 2023 Conference on Empirical Methods in Natural Language Processing}, pages 4268--4280, Singapore. Association for Computational Linguistics.

\bibitem[{Zhai et~al.(2023)Zhai, Mustafa, Kolesnikov, and Beyer}]{10377550}
Xiaohua Zhai, Basil Mustafa, Alexander Kolesnikov, and Lucas Beyer. 2023.
\newblock \href {https://doi.org/10.1109/ICCV51070.2023.01100} {Sigmoid loss for language image pre-training}.
\newblock In \emph{2023 IEEE/CVF International Conference on Computer Vision (ICCV)}, pages 11941--11952.

\bibitem[{Zhang et~al.(2023{\natexlab{a}})Zhang, Liu, Zhang, Sun, Xie, and Zha}]{10.1145/3581783.3612183}
Fanrui Zhang, Jiawei Liu, Qiang Zhang, Esther Sun, Jingyi Xie, and Zheng-Jun Zha. 2023{\natexlab{a}}.
\newblock \href {https://doi.org/10.1145/3581783.3612183} {Ecenet: Explainable and context-enhanced network for muti-modal fact verification}.
\newblock In \emph{Proceedings of the 31st ACM International Conference on Multimedia}, MM '23, page 1231–1240, New York, NY, USA. Association for Computing Machinery.

\bibitem[{Zhang et~al.(2020)Zhang, Kishore, Wu, Weinberger, and Artzi}]{bert-score}
Tianyi Zhang, Varsha Kishore, Felix Wu, Kilian~Q. Weinberger, and Yoav Artzi. 2020.
\newblock \href {https://openreview.net/forum?id=SkeHuCVFDr} {Bertscore: Evaluating text generation with {BERT}}.
\newblock In \emph{8th International Conference on Learning Representations, {ICLR} 2020, Addis Ababa, Ethiopia, April 26-30, 2020}. OpenReview.net.

\bibitem[{Zhang et~al.(2023{\natexlab{b}})Zhang, Trinh, Cao, Cui, and Liu}]{zhang2023interpretable}
Yizhou Zhang, Loc Trinh, Defu Cao, Zijun Cui, and Yan Liu. 2023{\natexlab{b}}.
\newblock \href {https://arxiv.org/abs/2304.07633} {Interpretable detection of out-of-context misinformation with neural-symbolic-enhanced large multimodal model}.
\newblock \emph{ArXiv preprint}, abs/2304.07633.

\end{thebibliography}

\appendix

\begin{table*}
    \centering
    \resizebox{\textwidth}{!}{ %
    \begin{tabular}{ccc}
    \hline
       Category  & Objects & Prompt \\
    \hline
      general & entire image &  Answer in one to three sentences: what are the people, objects, animals, \\
      &  & events, texts shown in the image? \\
      people & Person & Who is shown in the image?   \\
      animals   & Animal, Bird, Cat, Dog, Fish & Which \{\} species is shown in this image? \\
      buildings & Building, Stadium,  Bridge, Castle  &  Which \{\} is shown in this image? Provide a location if possible. \\
      flags & Flag & Which flag is shown in this image? \\
      food & Food, Drink, Fruit & Which \{\} is shown in this image? \\
      sports & Basketball,  Baseball bat  & What are the teams playing in this game? \\
       & Baseball glove, Football, Rugby ball & \\
       transports &  Airplane, Boat, Bus, Car, Helicopter &  Which \{\} model is shown in this image?\\
      & Motorcycle, Ship, Tank, Train, Truck, Van & \\
       weapons & Weapon & Which weapon model is shown in this image?\\
       \hline
    \end{tabular}}
    \caption{Parent categories of detected objects with their captioning prompts.}
    \label{tab:object_categories}
\end{table*}

\begin{figure*}
    \centering
    \begin{subfigure}{0.5\textwidth}
        \includegraphics[width=\textwidth]{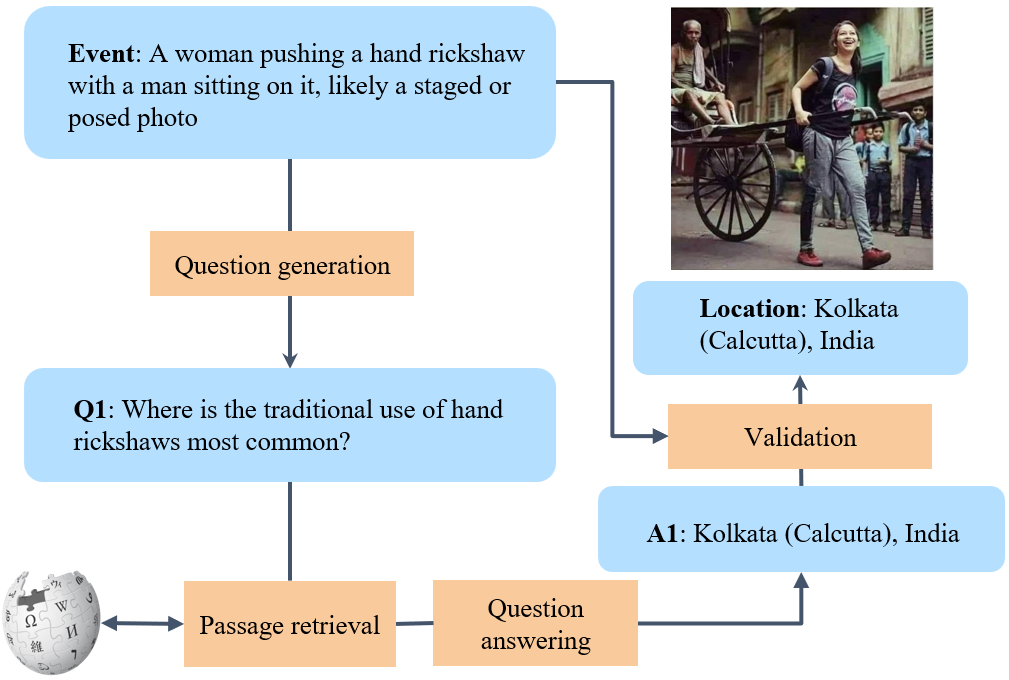}
        \caption{Missing location.}
        \label{fig:subfig_a}
    \end{subfigure}%
    \hfill
    \begin{subfigure}{0.5\textwidth}
        \includegraphics[width=\textwidth]{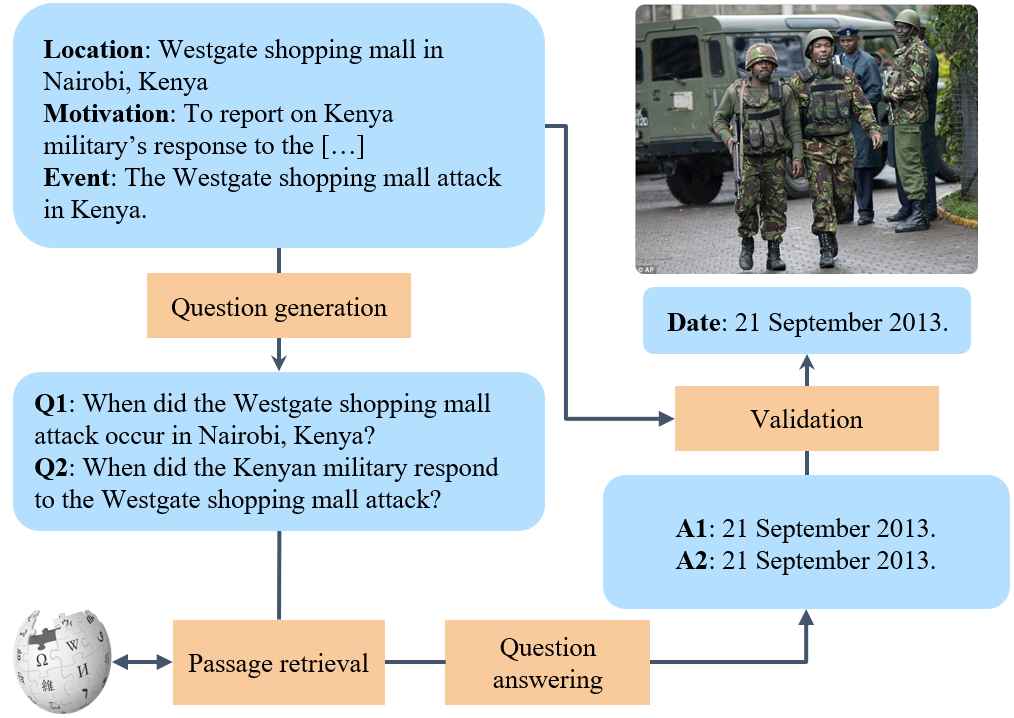}
        \caption{Missing date.}
        \label{fig:subfig_b}
    \end{subfigure}%
    \caption{Two examples of successful knowledge gap completion on 5Pils-OOC.}
    \label{fig:knowledge_completion_examples}
\end{figure*}
\section{Object categories for automated captions}
\label{sec:object_categories}

Table \ref{tab:object_categories} provides the mapping from the object labels of the Google Vision API to our manually defined parent categories. Each category is associated with one captioning prompt that is provided to LlavaNext to caption the cropped image of the detected object.

\section{Definition of the context items}
\label{sec:items}

We consider seven context items which we define below. They are paired with a question and a set of relevant named entities, which are used to rank the web captions.

\textit{Source} is the entity who created the image or first published it online \citep{tonglet-etal-2024-image}. The corresponding question is ``Who is the source of the image?''. The corresponding named entity is ``ORG''. 

\textit{Date} is the time at which the image was captured or created \citep{tonglet-etal-2024-image}. The corresponding question is ``When was the image taken?''. The corresponding named entity is ``DATE''. 

 \textit{Location} is the place where the image was captured \citep{tonglet-etal-2024-image}. The corresponding question is ``Where was the image taken?''. The corresponding named entities are ``FAC'', ``GPE'', ``LOC''. 

 \textit{Motivation} is the reason why the source took the image, usually to report on a news event, although other intents are possible \citep{tonglet-etal-2024-image}. The corresponding question is ``Why was the image taken?''. The corresponding named entities are ``EVENT'', ``GPE'', ``NORP'', ``ORG''.

 \textit{People} is the people that can be seen in the image. The corresponding question is ``Who is shown in the image?''. The corresponding named entity is ``PERSON''.
 
 \textit{Things} is a broad category that includes every entity that is shown in the image and not a human being. The corresponding question is ``Which animals, plants, buildings, or objects are shown in the image?''. The corresponding named entities are ``FAC'', ``LOC'', ``PRODUCT''.
 
 \textit{Event} describes the circumstances surrounding the image. The corresponding question is ``Which event is depicted in the image?''. The corresponding named entities are ``EVENT'', ``NORP''.
 
\citet{dufour2024ammeba} showed that the context items that are the most frequently altered when creating OOC captions are the \textit{date} and the \textit{event}, around 25\% each, followed by \textit{Location} around 18\%.  \textit{People} and \textit{things} reach together around 15\%.

\begin{figure*}   
\centering
\begin{tcolorbox}[colback=gray!5, colframe=gray!80, title=Caption decomposition prompt, width=\linewidth]
You are a helpful assistant who provides a structured summary of the content of an image.

The summary should take the form of a JSON file with the following key entries: “people” (the people shown in the image as a list), “object” (the objects shown in the image as a list), “event” (the event depicted in the image), “date” (the date when the image was taken), “location” (the location where the image was taken), “motivation” (the reason why the photographer took the image, such as for documenting, reporting, or for personal or organizational purposes. This reflects the photographer's intent. Do not assume it if not specified), “source” (the original author of the image, a person or an organization name). 

 You need to provide the answer to each of those items based on the caption of the image. Note that the caption does not always describe the image content. If there is no information in the caption for one of the key entries, you should write “not enough information” for that entry of the JSON file. Your answer should only contain the JSON file as a dictionary.
\end{tcolorbox}
    \caption{Prompt template for caption decomposition with Llama 3.}
\label{fig:decomp_prompt}
\end{figure*}

\begin{figure}
\centering
\begin{tcolorbox}[colback=gray!5, colframe=gray!80, title=Accurate caption generation prompt, width=\linewidth]
You are given a date, a location, and a motivation describing an image. Combine the 3 in one sentence of maximum 30 words.

Write the facts only, avoid journalistic style and adjectives, avoid introducing new information.

 Date : \{\} , Location : \{\}, Motivation : \{\}
\end{tcolorbox}
    \caption{Prompt to generate an accurate caption with GPT4, given the ground truth context items.}
\label{fig:accurate_caption}
\end{figure}

\begin{figure}
    \centering
    \includegraphics[width=\linewidth]{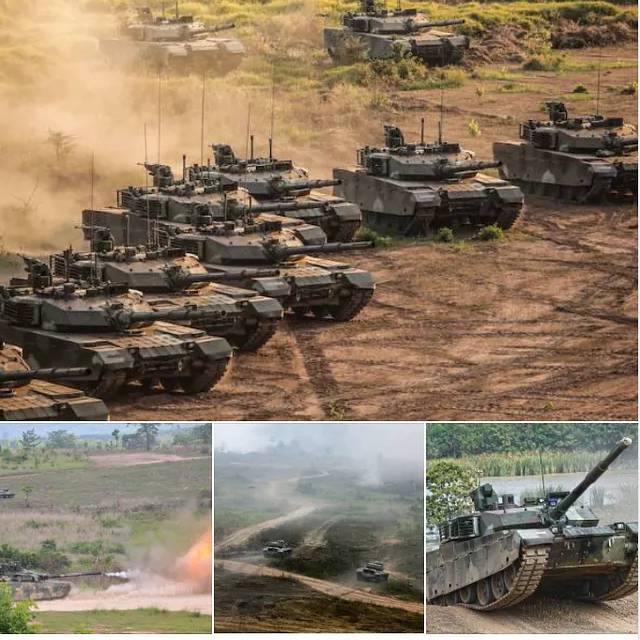}
    \caption{A composite image, where the sub-images have distinct true contexts.}
    \label{fig:composite_image}
\end{figure}

\section{Examples of knowledge gap completion}
\label{sec:knowledge_gap}

Figure \ref{fig:knowledge_completion_examples} shows two examples of knowledge gap completion on 5Pils-OOC. The question generation and the retrieval of a relevant Wikipedia passage allows to predict the location in the first image and the date in the second one.

\section{Prompts for caption decomposition}
\label{sec:GT_context}

To obtain the ground truth context items for NewsCLIPpings and for parts of 5Pils-OOC, we task Llama 3 to decompose the accurate caption of the image as a dictionary with context items as keys. Figure \ref{fig:decomp_prompt} shows the prompt.

\begin{table*}
    \centering
    \resizebox{\textwidth}{!}{ %
    \begin{tabular}{cccccccccccc}
    \hline
         & \multicolumn{2}{c}{Source} & \multicolumn{2}{c}{Date}  & \multicolumn{4}{c}{Location} & \multicolumn{3}{c}{Motivation}\\
         & (RL) & (M)  & (EM) & ($\Delta$) & (RL) & (M) &  (CO$\Delta$) &  (HL$\Delta$)  & (RL) & (M) & (BertS)  \\
    \hline 
    & \multicolumn{11}{c}{\textbf{NewsCLIPpings}} \\
    \hline
    
       5Pils baseline & 5.0 & 3.4 & 23.1 & 28.9 & 34.4 & 32.7 & 38.7 & 33.1 & 1.0  & 1.5 & 9.6\\
             \textbf{\textcolor{myblue}{CO}\textcolor{myorange}{VE}}   & \textbf{12.2} & \textbf{8.1} & \textbf{30.7} & \textbf{41.5} & \textbf{44.2} & \textbf{42.4} & \textbf{51.6} & \textbf{43.1} & \textbf{5.6}  &  \textbf{11.6} & \textbf{65.3} \\
    \hline
        & \multicolumn{11}{c}{\textbf{5Pils-OOC}}\\
    \hline
      5Pils baseline & 0.4 & 0.3 & 0.5 & 1.8 & 15.2 & 12.1 & 21.8 & 16.8 & 3.0 & 3.0 & \textbf{62.0} \\
     \textbf{\textcolor{myblue}{CO}\textcolor{myorange}{VE}}   & \textbf{0.9} & \textbf{0.6} & \textbf{1.1} & \textbf{7.0} & \textbf{18.6} & \textbf{16.7} & \textbf{28.9} & \textbf{22.5} & \textbf{17.1} & \textbf{15.1} & 56.1 \\
    \hline
    \vspace{3pt}\\
    \hline
     & \multicolumn{3}{c}{People} & \multicolumn{3}{c}{Things} & \multicolumn{3}{c}{Event} & &  \\
      & (R) & (P) & (F1) & (RL) & (M) & (BertS)  & (RL) & (M) & (BertS) & & \\ 
  \hline 
    & \multicolumn{11}{c}{\textbf{NewsCLIPpings}} \\
    \hline

      5Pils baseline & 38.9 & 42.6 & 39.9 & \textbf{12.2} & 9.9 & 62.6  &  14.1 & 15.4 & 43.6 & &  \\
     \textbf{\textcolor{myblue}{CO}\textcolor{myorange}{VE}}   & \textbf{48.2} & \textbf{51.5} & \textbf{49.0} & 9.7 & \textbf{10.9} & \textbf{63.0}  & \textbf{14.6} & \textbf{22.4} & \textbf{68.1} & & \\
    \hline
        & \multicolumn{11}{c}{\textbf{5Pils-OOC}} \\
    \hline
   5Pils baseline & 12.1 & 14.1 &  12.8 & 6.9 & 4.9 & \textbf{63.9} & 5.3 &  4.8 & \textbf{65.0} &   &  \\
     \textbf{\textcolor{myblue}{CO}\textcolor{myorange}{VE}}   & \textbf{20.8} & \textbf{21.5} &  \textbf{20.5} & \textbf{6.8} & \textbf{7.2} & 61.2 & \textbf{7.0} & \textbf{9.4} & 51.7 & &  \\
    \hline     
    \end{tabular}}
    \caption{Detailed context prediction results on the test sets (\%). Best results are marked in \textbf{bold}.}
    \label{tab:context_detailed_results}
\end{table*}

\section{Creation of 5Pils-OOC}
\label{sec:5Pils_OOC}

5Pils \citep{tonglet-etal-2024-image} is the first real-world misinformation dataset that provides labels for context items. The ground truth context of the image is obtained by extracting the context items from an FC article written by human experts. To use 5Pils for context and veracity prediction, we need to make some adjustments.

First, the scope of 5Pils is broader than ours. It includes other types of misinformation than OOC images, namely, manipulated and fake images. Hence, we start by removing all images that are manipulated or fake using the  ``type of image'' metadata field. Furthermore, 5Pils does not provide accurate captions. We generate accurate captions with GPT4 by combining the ground truth context items. This requires at least the \textit{motivation} item and one or two out of \textit{date} and \textit{location}. Images that do not satisfy those label criteria are discarded. Figure \ref{fig:accurate_caption} shows the prompt to generate accurate captions. Finally, some images in 5Pils are composite, that is, they are a collage of more than one image next to each other. An example is shown in Figure \ref{fig:composite_image}. Such images are considered out-of-scope.

\begin{figure*}
    \centering
\begin{tcolorbox}[colback=gray!5, colframe=gray!80, title=Context prediction prompt, width=\linewidth]
You are a helpful assistant who answers questions about an image based on captions from the web, automatically generated captions, and visual entities.\\

\noindent You are given captions from web pages that use the image.

You are also given two automatically generated captions. 
The first caption provides a global description of the image, sometimes including people but without naming them. 

The second caption provides more details about the people in the image, including their names and short biographies when that information is available.

You are also given visual entities which may be relevant to the image without certitude.\\

\noindent Here are a few examples:

[Demonstrations]\\ 

\noindent Web captions that may be relevant: \{\}

Caption 1 - Global description: \{\}

Caption 2 - More details about the people in the image: \{\}

Visual entities that may be relevant, without certitude: \{\}

Question: \{\} 

Answer in one sentence. Answer with one word ("unknown") if the information is not provided.

Answer: \{\}
\end{tcolorbox}
    \caption{Prompt template for context prediction with Llama 3.}
\label{fig:context_prompt}
\end{figure*}

\section{Detailed context prediction evaluation}
\label{sec:metrics}

Table \ref{tab:context_detailed_results} complements the results of Table \ref{tab:context_results} by showing additional metrics for the context items defined in \citet{tonglet-etal-2024-image}. This means computing  the RougeL (RL) score \citep{lin-2004-rouge}  for \textit{source}, the exact match (EM) for \textit{date}, the RougeL and Meteor scores for \textit{location}, and the RougeL and BertScore (BertS) \citep{bert-score} for \textit{motivation}, 
Furthermore, we compute HL$\Delta$ for \textit{location}, which is inversely proportional to the hierarchical distance in the GeoNames ontology between the prediction and ground truth. We also compute additional metrics for the new items introduced in this work. For \textit{people}, we compute the Recall (R) and Precision (P). For \textit{things} and \textit{event}, we add the RougeL and Bert scores. Consistent with our observations in Table \ref{tab:context_results}, we observe that COVE outperforms the baseline \citep{tonglet-etal-2024-image} for most metrics.

\begin{figure*}
    \centering
\begin{tcolorbox}[colback=gray!5, colframe=gray!80, title=Question generation prompt, width=\linewidth]
You are a helpful assistant. You are given context information about an image. The date of the image is unknown.

Your task is to generate one to three world knowledge questions that will help determine when the image was taken. The questions should either start with “When” or “On which date”. The question should be self-contained and specific enough to be answerable based on world knowledge.

If no self-contained and specific question can be generated, your response should be “No questions can be generated given the context”.\\ 

\noindent You are provided with examples below.

[Demonstrations]\\

\noindent Context information: \{\}

Generated questions (up to 3): \{\}
\end{tcolorbox}
    \caption{Prompt template for knowledge gap completion - question generation with Llama 3.}
\label{fig:qg_prompt}
\end{figure*}

\begin{figure*}
    \centering
\begin{tcolorbox}[colback=gray!5, colframe=gray!80, title=Question answering prompt, width=\linewidth]
You are a helpful assistant. You are given a question that requires world knowledge to be answered.

Your task is to provide a specific answer to the question in 1 or 2 sentences based on available knowledge from Wikipedia. Your answer should be a date at the day, month or year level. 

If the question cannot be answered based on the available knowledge, your response should be “Unknown”.\\

\noindent You are provided with examples below.

[Demonstrations]\\

\noindent Wikipedia knowledge:  \{\}

Question: \{\}

 Answer: \{\}

\end{tcolorbox}
    \caption{Prompt template for knowledge gap completion - question answering with Llama 3.}
\label{fig:qa_prompt}
\end{figure*}

\begin{figure*}
    \centering
\begin{tcolorbox}[colback=gray!5, colframe=gray!80, title=Validation prompt, width=\linewidth]
You are a helpful assistant. You are given context information about an image, as well as world knowledge information. 

Your task is to estimate when the image was taken or provide a plausible time range using the context and world knowledge information.

If the date cannot be derived from the context and world knowledge, your response should be “Unknown”. 

\noindent You are provided with examples below.

[Demonstrations]\\

\noindent Context: \{\}

World knowledge: \{\}

When was the image taken?

Answer: \{\}

\end{tcolorbox}
    \caption{Prompt template for knowledge gap completion - validation with Llama 3.}
\label{fig:validation_prompt}
\end{figure*}

\begin{figure*}
    \centering
\begin{tcolorbox}[colback=gray!5, colframe=gray!80, title=Veracity prediction prompt, width=\linewidth]
You are a helpful assistant that verifies if a caption is accurate for an image or if it is an attempt at out-of-context misinformation.
Instead of the image itself, you are given structured context information about the image. You need to verify whether the caption is supported by the context information.\\

\noindent Here are a few examples:

[Demonstrations]\\

\noindent Context information: \{\}

\noindent Caption to verify: \{\} 

Given the context information, is the caption accurate or is it out-of-context? 

If there are too many unknown elements to provide a clear decision, you can answer that the accuracy of the caption is “unknown”, potentially leaning more towards accurate or out-of-context.

Provide a detailed reasoning. Then provide your answer strictly among the following choices: “accurate”, “unknown, probably accurate”, “unknown”, “unknown, probably out-of-context”, “out-of-context”.\\

\noindent Reasoning: \{\}

Answer: \{\}

\end{tcolorbox}
    \caption{Prompt template for veracity prediction with Llama 3.}
\label{fig:veracity_prompt}
\end{figure*}

\section{Hyperparameters and model versions}
\label{sec:implementation}

\textbf{Web captions collection} We set  $t_{match}$   to 0.92 and  $t_{non\_match}$  to 0.7 for web images. Visual entities are included if their score is at least 0.1.

 \textbf{Wikipedia entities collection} The following types of named entities are extracted from the caption: ``PERSON'', ``FAC'', ``PRODUCT''. We set $k$, the number of nearest entities to retrieve from the OVEN index \citep{hu2023open} to 5. We set $t_{wiki\_text}$ to 0.23. If the entity  is a ``PERSON'', $t_{wiki\_image}$ is set to 0.92. Otherwise, it is set to 0.7. 

 \textbf{Automated captions generation} We keep \textit{Person} objects detected  with confidence scores above 0.8.

\textbf{Context prediction} We set $l$, the number of web captions to provide as input, to 10.

\textbf{Knowledge gap completion} For each question, we retrieve one Wikipedia passage, if its relevance score \citep{semnani-etal-2023-wikichat} is above 20.

\textbf{Model and versions}
We use the HuggingFace's transformers \citep{wolf-etal-2020-transformers} and vLLM \citep{10.1145/3600006.3613165} libraries to load, train, and make inferences with models. Named entities are detected using the Spacy model \textit{en\_web\_core\_lg}. We create the OVEN index with the FAISS library \citep{douze2024faiss}. We use the CLIP-ViT-L14 (\textit{openai/clip-vit-large-patch14}) version of CLIP. We also considered CLIP-ViT-B32 (\textit{openai/clip-vit-base-patch32}) and SIGLIP (\textit{google/siglip-so400m-patch14-384}) \citep{10377550}. On the validation set of NewsCLIPpings, both SIGLIP and CLIP-ViT-L14 achieve the same accuracy, but CLIP-ViT-L14 has a slightly higher R$_{OOC}$. Accuracy with CLIP-ViT-B32 is 1 percentage point lower. To compute similarities between the image and Wikipedia images for ``PERSON'' entities, we use the face-net library.\footnote{\href{https://pypi.org/project/face-recognition/}{pypi.org/project/face-recognition/}} For COVE, SNIFFER, and the baseline of \citet{tonglet-etal-2024-image}, we use the \textit{meta-llama/Meta-Llama-3-8B-Instruct}, \textit{llava-hf/llava-v1.6-mistral-7b-hf}, and \textit{microsoft/deberta-v3-large} versions of Llama 3, LlavaNext, and DebertaV3, respectively.  We set the temperatures to 0 for reproducibility. We fine-tune  DebertaV3  for 5 epochs, using a batch size of 4, a weight decay of 0.01, and a learning rate of 5e-6. All experiments are conducted with one A100 GPU.

\begin{figure*}
    \centering
    \includegraphics[width=\linewidth]{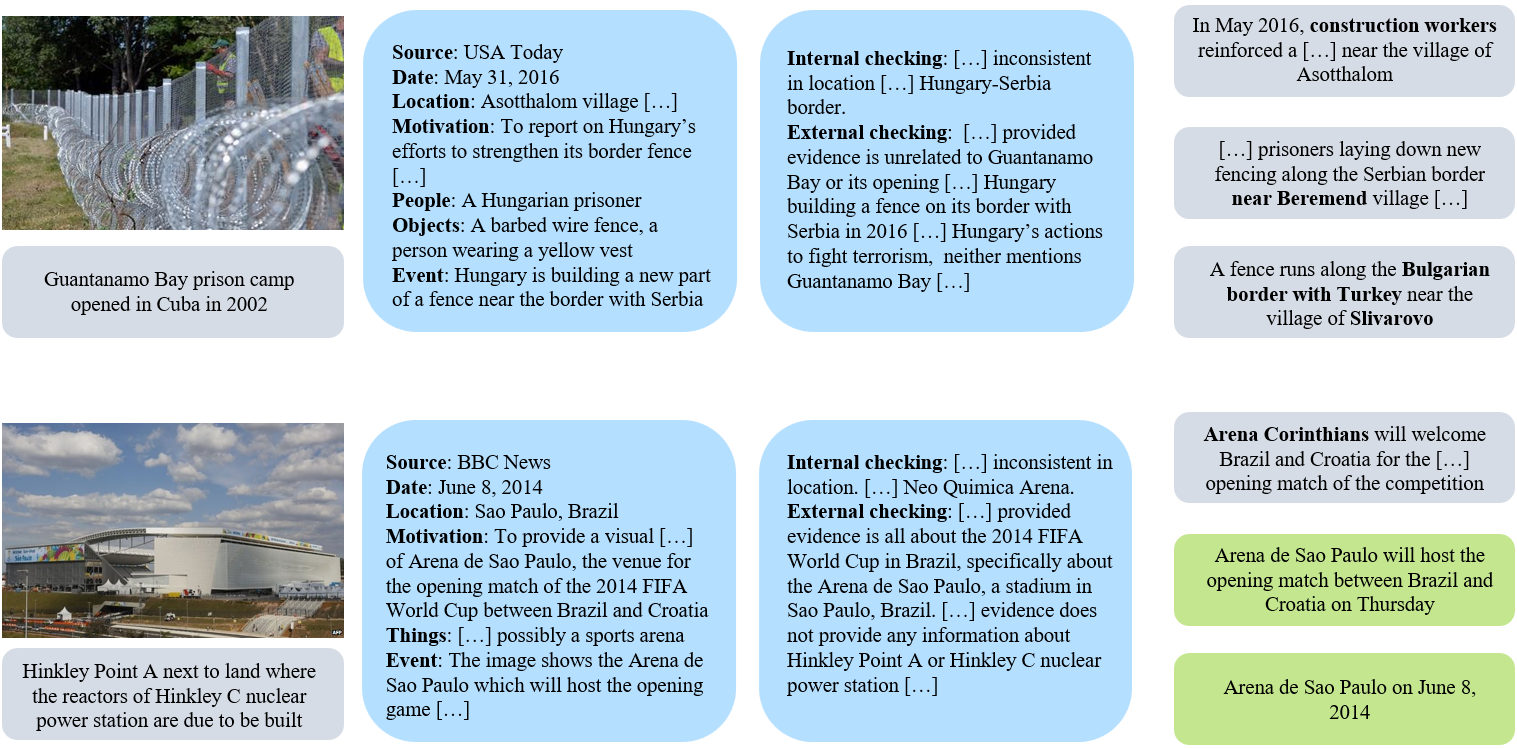}
    \caption{Human study examples. From left to right, the image with the old caption, the COVE artifact, the SNIFFER artifact, the three new captions. \colorbox{mylightgreen}{Accurate} captions are shown in green, \colorbox{mygrey}{OOC} captions in grey.}
    \label{fig:human_study}
\end{figure*}

\section{COVE - Llama 3 prompt templates}
\label{sec:prompts}

Figure \ref{fig:context_prompt} to \ref{fig:veracity_prompt} show the prompts used for context prediction, knowledge gap completion, and veracity prediction with Llama 3.
Each prompt starts with a task description, followed by a set of 4 to 8 demonstrations. For veracity predictions, there are two sets of demonstrations, one for instances with web captions and one for those without web captions. The template shows the structure of the demonstrations and the test instance. \{\} are replaced by the value of the demonstrations or the test instance. For knowledge gap completion, we provide the prompts used for \textit{date} prediction.

\section{Human study corpus creation}
\label{sec:human_statistics}

The human study corpus contains 30 images with OOC captions sampled from the NewsCLIPpings test set. We restrict the selection to images with at least six predicted context items. 
For each image, we create three new captions. This results in a corpus of 90 new captions, of which 22 are accurate and 68 OOC. The number of accurate captions per image ranges from 0 to 2.

Accurate captions are either the original caption from Visual News \citep{liu-etal-2021-visual} or a hand-written paraphrase based on additional context from the source news article.

Parts of the OOC captions are sampled from other splits of NewsCLIPpings that contain the same image. Others are written by hand by paraphrasing the original caption from Visual News and altering key context items based on the source news article, following the OOC misinformation techniques detailed in \citet{dufour2024ammeba}.

The low accuracy of the annotators in the absence of AFC artifacts and the moderate increase given the artifacts confirm the challenging nature of the created corpus.

\begin{figure*}
    \centering
    \includegraphics[width=\linewidth]{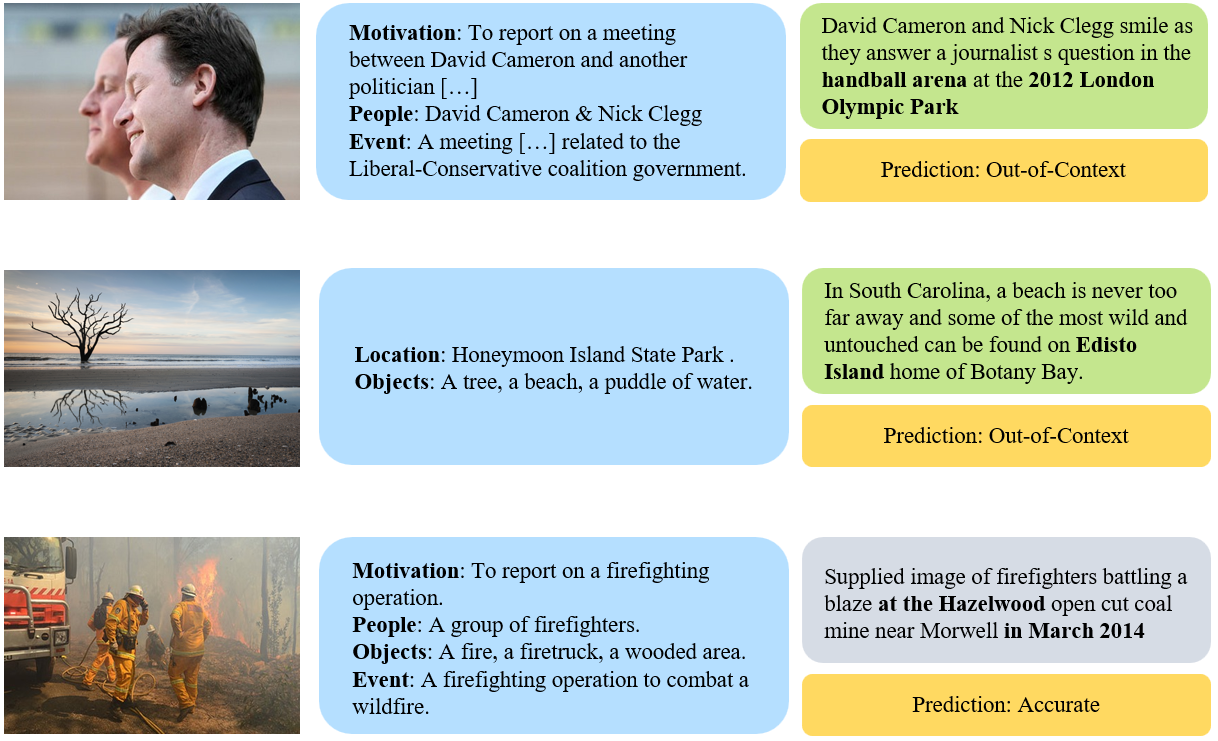}
    \caption{Error examples with Llama 3. Each row shows the image (left), the context prediction (center), the caption (upper right), and the veracity prediction (lower right).  \colorbox{mylightgreen}{Accurate} captions are shown in green, \colorbox{mygrey}{OOC} captions in grey.}
    \label{fig:qualitative}
\end{figure*}

\section{Human study instructions and examples}
\label{sec:human_example}

\begin{table}
    \centering
    \resizebox{\linewidth}{!}{ %
    \begin{tabular}{cccc}
    \hline
         & Not confident & Somewhat confident & Highly confident \\
    \hline
       Group 1 - no AFC artifact  & 45.9 & 37.8 & 16.3    \\
       Group 1 - SNIFFER artifact & 14.8 & 54.4 & 30.7 \\
       Group 2 - no AFC artifact  & 38.5 & 48.9 & 12.6\\
       Group 2 - COVE artifact & 8.1 & 33.0 & 58.9 \\
    \hline
    \end{tabular}}
    \caption{Confidence levels of the annnotators (\%) in different setups. Each row sums to 100\%.}
    \label{tab:confidence}
\end{table}

The following instructions were provided to the participants in the first phase: ``Your task is to decide if a caption is accurate for an image or if it is out-of-context.
Out-of-context means that the caption is (partially or totally) not matching the image (in terms of people, event, date, location, objects, ...). The caption may or may not describe a real event, but it is not accurate for the given image.
You will be given an image and a caption that has already been assessed as out-of-context, based on that you need to classify 3 new captions as Accurate or out-of-context, and indicate your confidence level.
Important : do not search for information online or conduct reverse image search with the image. You are expected to answer based only on the image and the previously fact-checked caption.''

Afterward, the participants are given the instructions for the second phase. If they are given the COVE artifact, they receive the following instructions: ``The task is the same, and you will see the same images and captions to verify
but this time, you are also given a summary of the image context, generated with the AFC method COVE.
You can now update your verdict and confidence scores based on this additional input.
''. If they are given the SNIFFER artifact,  they receive these instructions: ``The task is the same, and you will see the same images and captions to verify
but this time, you are also given the explanations generated with the AFC method SNIFFER. SNIFFER explanations consist of 2 parts: Internal checking: the model compared the image with the previously fact-checked caption, as well as a set of relevant visual entities, and detected inconsistencies if any. External checking: the model compared the previously fact-checked caption with relevant web evidence found by doing a reverse image search. You can now update your verdict and confidence scores based on this additional input.''

Figure \ref{fig:human_study} shows two examples of the corpus.

\section{Human study confidence levels}
\label{sec:confidence_level}

We report in Table \ref{tab:confidence} the non-aggregated confidence levels of the participants. For both groups, the  ``Highly confident'' level is not frequent during the first phase. After seeing the artifact, the most frequent level becomes ``Somewhat confident'' for SNIFFER and  ``highly confident'' for COVE.

\section{Error examples with COVE - Llama 3}
\label{sec:qualitative_analysis}

Figure \ref{fig:qualitative} provides three error examples of Llama 3 on the NewsCLIPpings test set. In the first example, the predicted context is missing the \textit{date} and \textit{location} items. Llama 3 wrongly predicts the caption as OOC, given that not all atomic facts are supported by the context. The second image is an example of context prediction error propagating to veracity prediction. While Edisto Island is the correct location of the image, the predicted \textit{location} is incorrect. As a result, the caption is wrongly predicted as OOC. The third example is the opposite of the first example. The predicted context is not sufficient to verify all atomic facts, but the caption is wrongly predicted as accurate.

\section{Filtering mechanism experiment}
\label{sec:filtering}

We conduct an experiment on 5Pils-OOC where we only use web evidence that belongs to a manually curated list of trustworthy sources. The sources are: theguardian.com, usatoday.com, nytimes.com, washingtonpost.com, reuters.com, indiatimes.com, bbc.com, cnn.com, nbcnews.com, thetimes.co.uk, and apnews.com. By selecting web evidence from these sources only, the accuracy decreases by 2.3 and the Macro F1 score by 2.8 percentage points. Only R$_{OOC}$ decreases while R$_{ACC}$ remains unchanged. Despite the simplicity of this filtering approach, the decrease in veracity prediction performance is relatively small, indicating that a more advanced filtering mechanism could provide results equivalent to an approach without filtering or even outperform it.

\end{document}